\documentclass[10pt,twocolumn,letterpaper]{article}

\usepackage{cvpr}              

\usepackage[dvipsnames]{xcolor}

\definecolor{cvprblue}{rgb}{0.21,0.49,0.74}
\usepackage[pagebackref,breaklinks,colorlinks,citecolor=cvprblue]{hyperref}

\usepackage{amsmath,amsfonts}
\usepackage{algorithmic}
\usepackage{algorithm}
\usepackage{array}
\usepackage{textcomp}
\usepackage{stfloats}
\usepackage{url}
\usepackage{verbatim}
\usepackage{graphicx}
\usepackage{cite}
\usepackage{graphicx}
\usepackage{amsmath}
\usepackage{amssymb}
\usepackage{booktabs}

\usepackage{multirow}
\usepackage{algorithm}
\usepackage{algorithmic}

\usepackage{setspace}

\usepackage{colortbl}
\usepackage{xcolor}
\usepackage{array} 
\usepackage{float}
\usepackage{arydshln}

\usepackage{svg}

\usepackage{setspace}

\def\paperID{8025} 
\def\confName{CVPR}
\def\confYear{2024}

\title{A Semi-supervised Nighttime Dehazing Baseline with Spatial-Frequency Aware and Realistic Brightness Constraint}

\author{Xiaofeng Cong\textsuperscript{\rm 1} \hspace{0.2cm} Jie Gui\textsuperscript{\rm 1}\thanks{Corresponding author} \hspace{0.2cm} Jing Zhang\textsuperscript{\rm 2} \hspace{0.2cm} Junming Hou\textsuperscript{\rm 1} \hspace{0.2cm} Hao Shen\textsuperscript{\rm 3}\\
\textsuperscript{\rm 1}Southeast University \hspace{0.2cm} \textsuperscript{\rm 2}University of Sydney \hspace{0.2cm} \textsuperscript{\rm 3}Hefei University of Technology\\
{\tt\small cxf\_svip@163.com, \{guijie,junming\_hou\}@seu.edu.cn, \{jingzhang.cv,haoshenhs\}@gmail.com}}

\begin{document}
\maketitle

\begin{abstract}
  Existing research based on deep learning has extensively explored the problem of daytime image dehazing. However, few studies have considered the characteristics of nighttime hazy scenes. There are two distinctions between nighttime and daytime haze. First, there may be multiple active colored light sources with lower illumination intensity in nighttime scenes, which may cause haze, glow and noise with localized, coupled and frequency inconsistent characteristics. Second, due to the domain discrepancy between simulated and real-world data, unrealistic brightness may occur when applying a dehazing model trained on simulated data to real-world data. To address the above two issues, we propose a semi-supervised model for real-world nighttime dehazing. First, the spatial attention and frequency spectrum filtering are implemented as a spatial-frequency domain information interaction module to handle the first issue. Second, a pseudo-label-based retraining strategy and a local window-based brightness loss for semi-supervised training process is designed to suppress haze and glow while achieving realistic brightness. Experiments on public benchmarks validate the effectiveness of the proposed method and its superiority over state-of-the-art methods. The source code and Supplementary Materials are placed in the https://github.com/Xiaofeng-life/SFSNiD.
\end{abstract}

\section{Introduction}
\label{sec:intro}

\begin{figure}
  \centering
  \scriptsize
  \includegraphics[width=2.7cm,height=1.7cm]{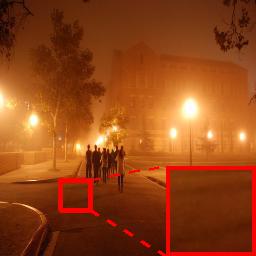}
  \includegraphics[width=2.7cm,height=1.7cm]{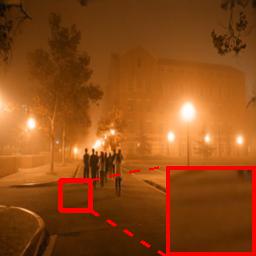}
  \includegraphics[width=2.7cm,height=1.7cm]{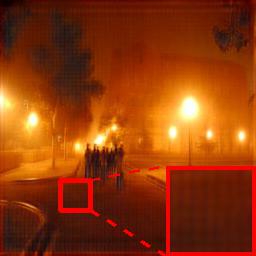}
  \\
  \includegraphics[width=2.7cm,height=1.7cm]{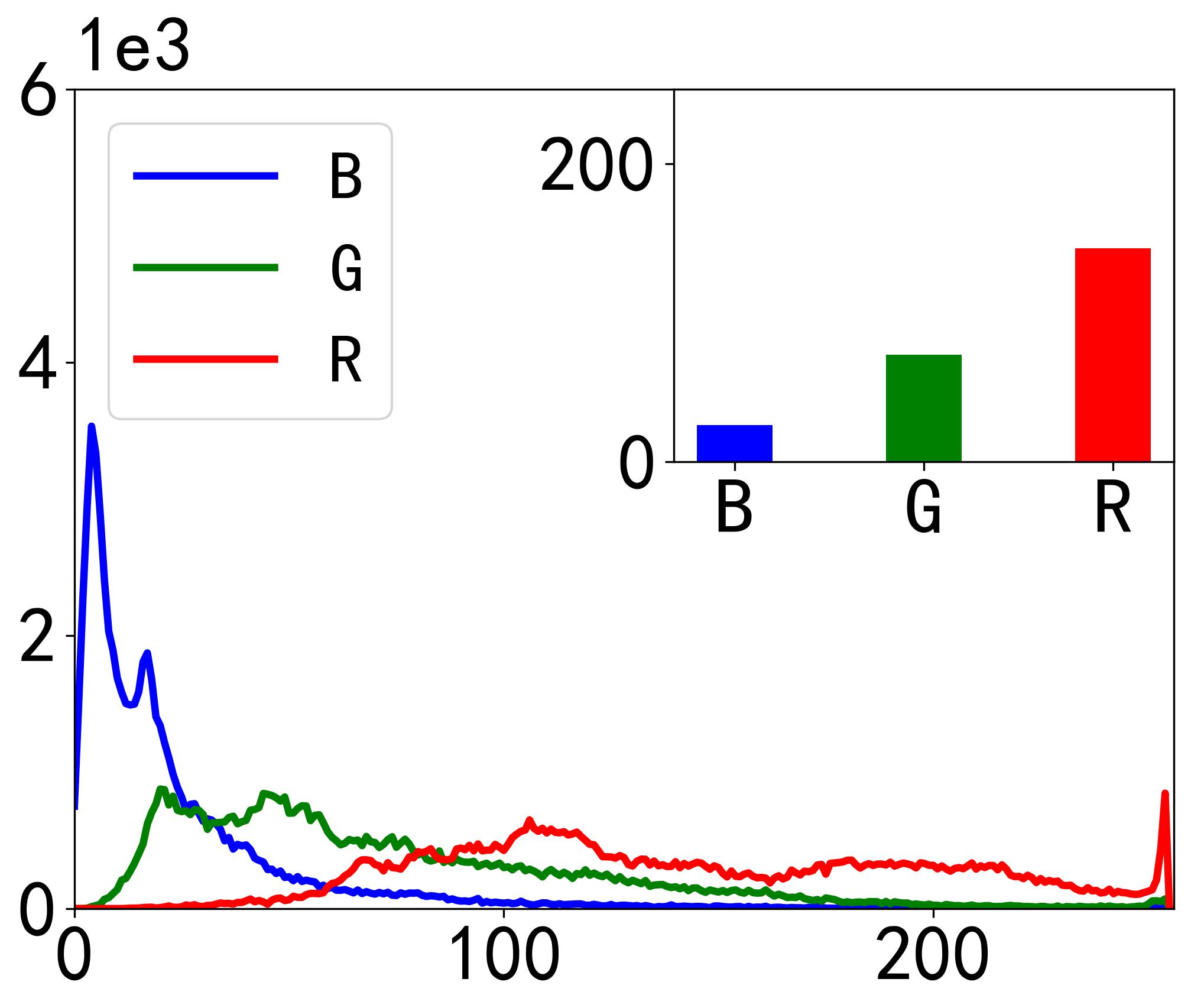}
  \includegraphics[width=2.7cm,height=1.7cm]{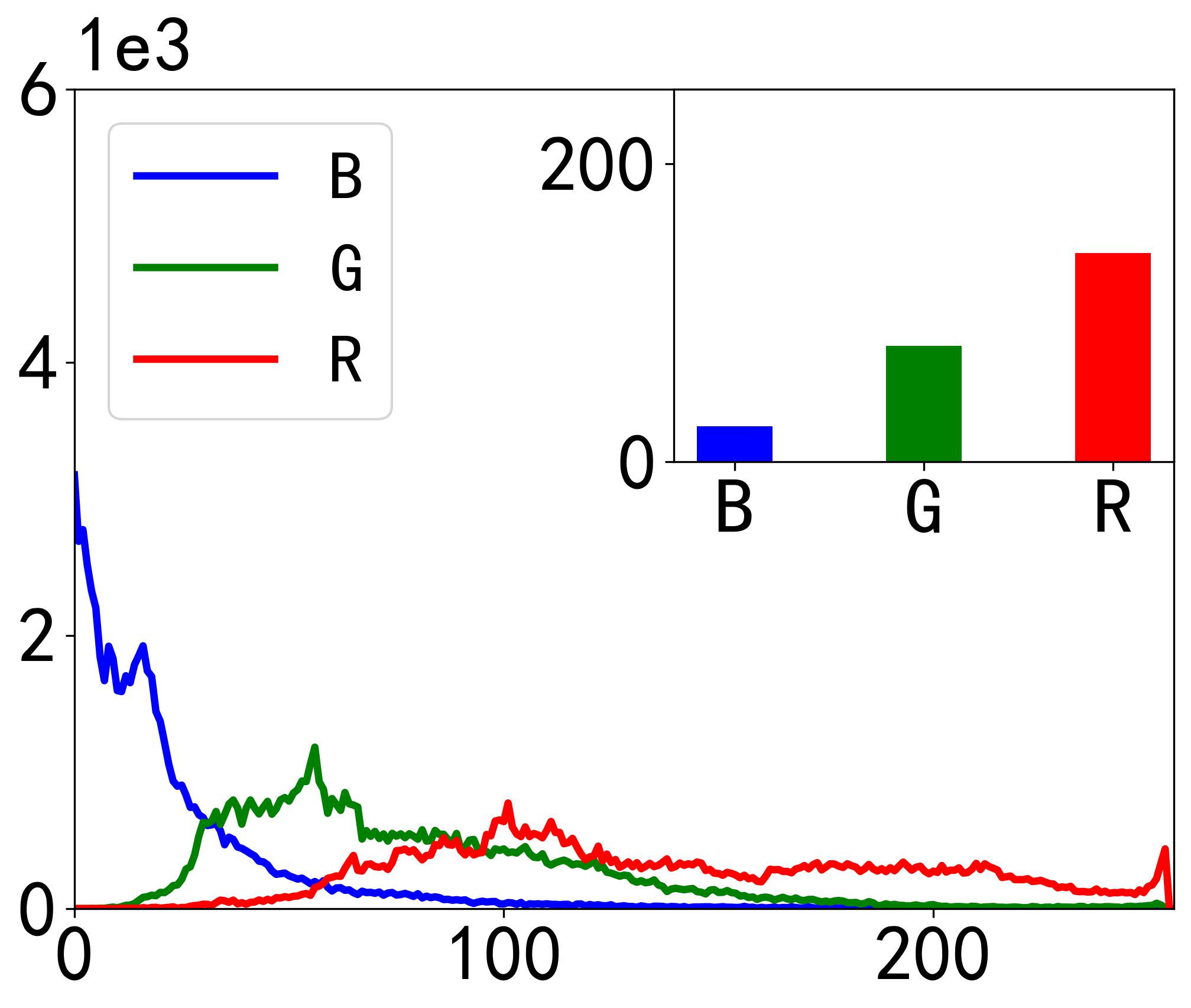}
  \includegraphics[width=2.7cm,height=1.7cm]{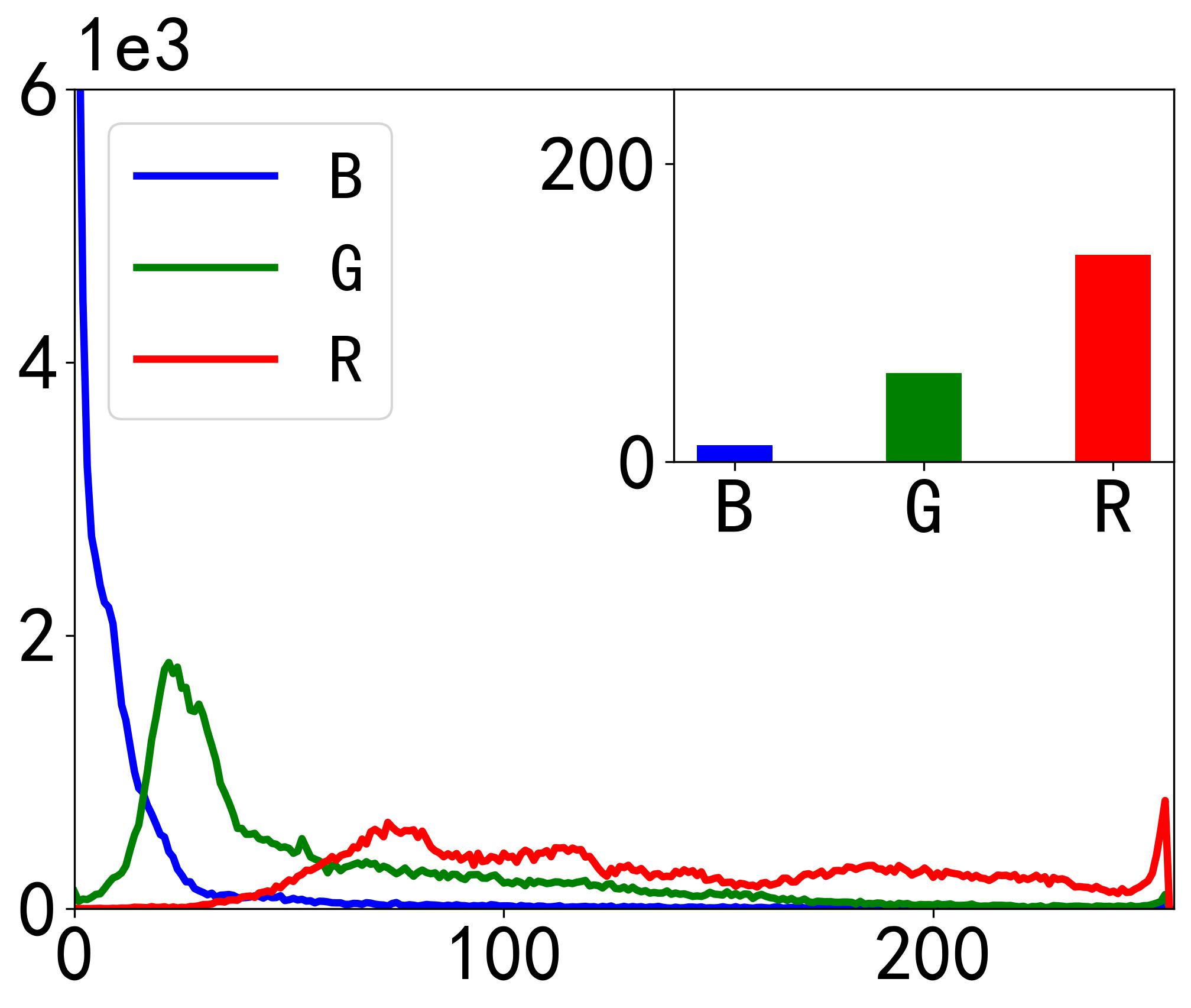}
  \\
  \leftline{\hspace{0.8cm} (a) Hazy \hspace{1.1cm} (b) IM-YellowHaze \cite{liao2018hdp} \hspace{0.4cm} (c) IM-NightHaze \cite{liao2018hdp}}
  \vspace{0.1cm}
  \includegraphics[width=2.7cm,height=1.7cm]{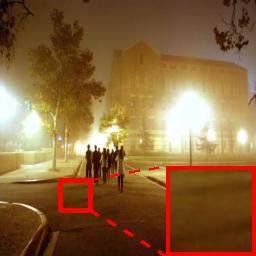}
  \includegraphics[width=2.7cm,height=1.7cm]{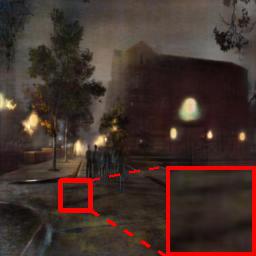}
  \includegraphics[width=2.7cm,height=1.7cm]{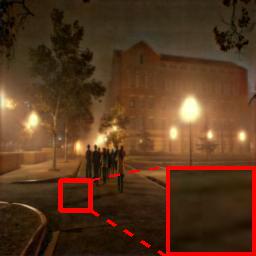}
  \\
  \includegraphics[width=2.7cm,height=1.7cm]{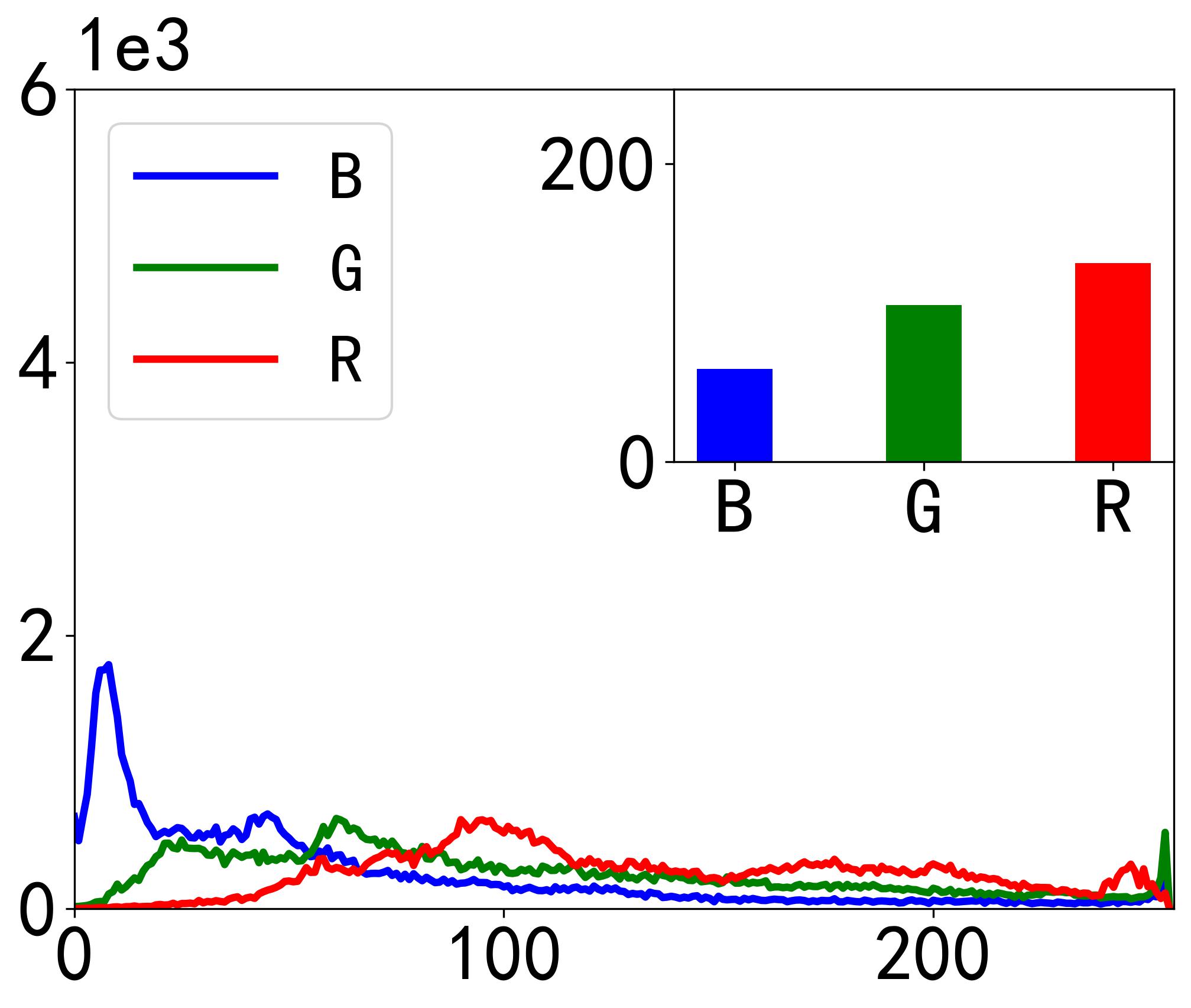}
  \includegraphics[width=2.7cm,height=1.7cm]{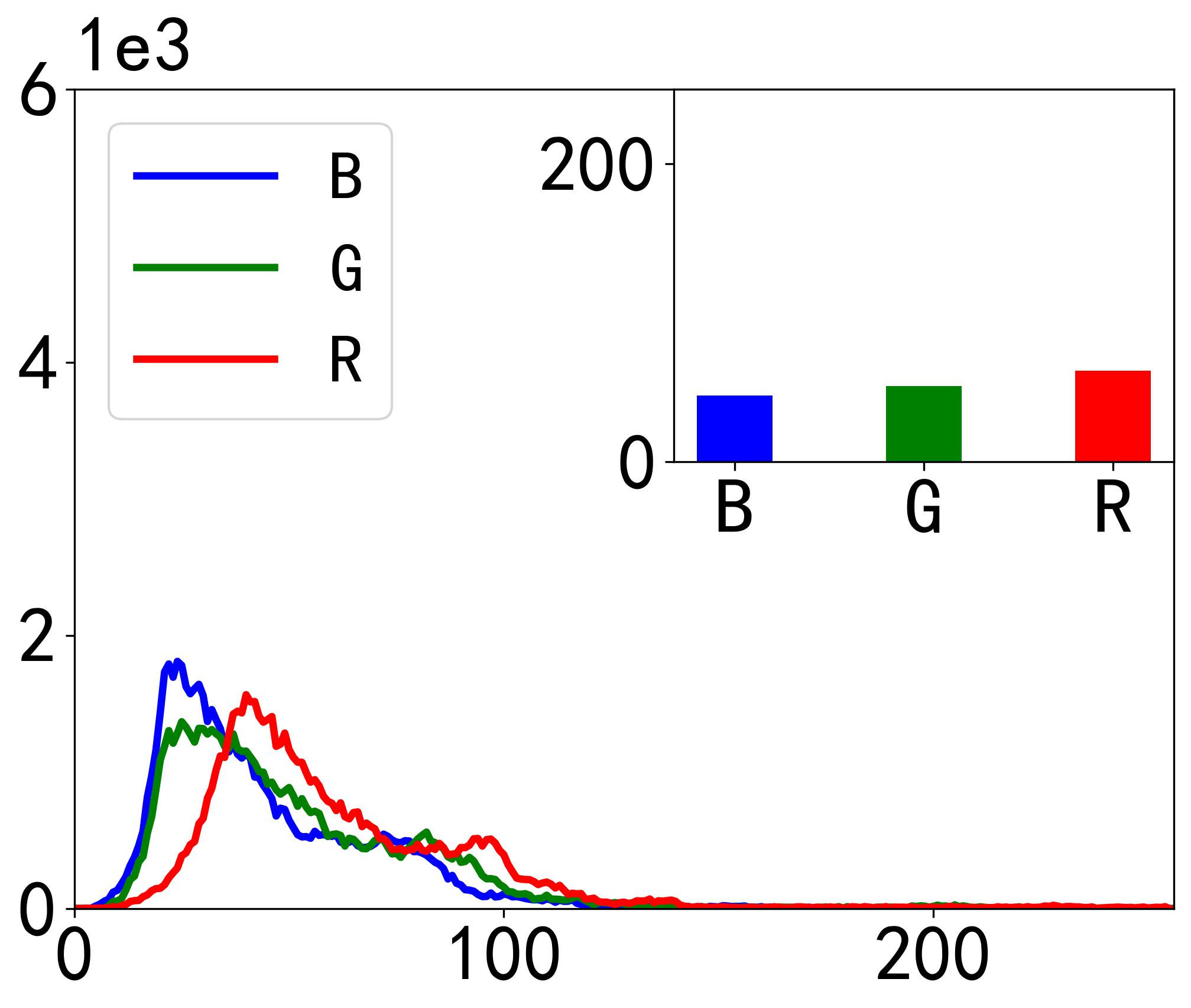}
  \includegraphics[width=2.7cm,height=1.7cm]{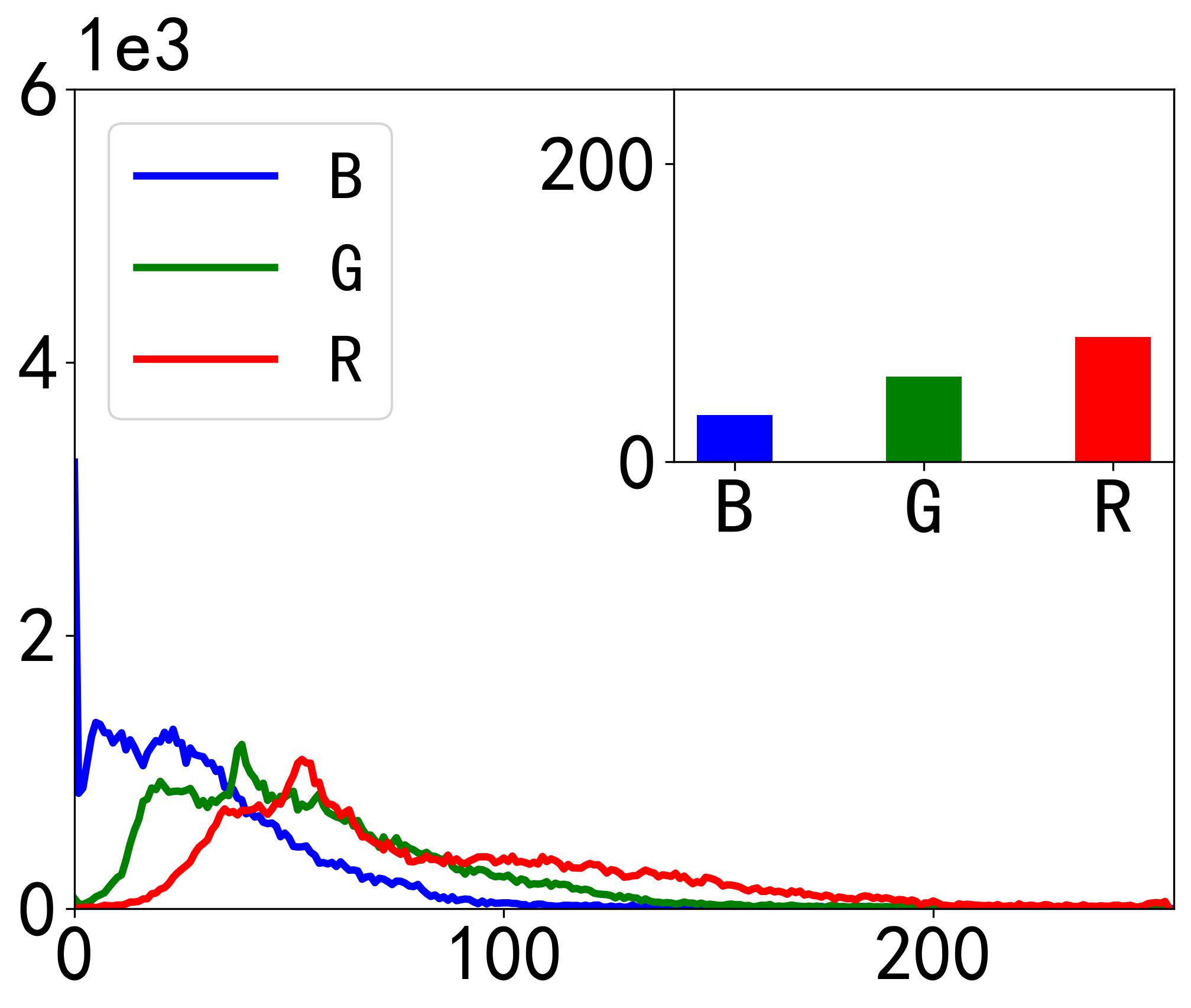}
  \leftline{\hspace{0.5cm} (d) IM-NHR \cite{zhang2020nighttime} \hspace{0.5cm} (e) GE-UNREAL-NH \cite{liu2023nighthazeformer} \hspace{0.7cm} (f) Ours}
  
  \caption{Visualization of real-world dehazed images, where the ``IM-'' and ``GE-'' denote the dehazed results obtained by training on  imaging model (IM) and game engine (GE) simulated datasets, respectively. The curve figure represents the pixel histogram, where the $x$ and $y$ coordinates represent the pixel values and corresponding numbers, respectively. The $x$ and $y$ coordinates of the bar figure represent the color channel and the corresponding average pixel value, respectively.
  }
  \label{fig:visual_results_of_histogram_hazy_and_dehazed}
\end{figure}

Nighttime and daytime images may contain hazy effects, which may cause their quality to be degraded \cite{zhang2017fast,gui2023a,sun2022rethinking,ju2021idrlp}. Therefore, two valuable research fields are proposed, which are daytime single image dehazing (DaSID) \cite{song2023vision,cong2020discrete,zheng2021ultra} and nighttime single image dehazing (NiSID) \cite{liu2023nighthazeformer,jin2023enhancing,kuanar2022multi}, respectively. Compared with the daytime hazy image, the imaging of the nighttime hazy image is more complex \cite{zhang2014nighttime,liu2021single}. Currently, NiSID is still a challenging problem. 

Existing research on DaSID \cite{wu2021contrastive,liu2019griddehazenet,zhang2019famed,ren2020single,liang2022self,zhang2021semantic,li2017aod,zhang2021hierarchical} have achieved impressive performance. Various effective DaSID algorithms have been proposed and verified on benchmark daytime datasets \cite{li2018benchmarking}. However, these DaSID algorithms are designed for the properties of daytime hazy and haze-free images, without taking into account the characteristics of nighttime hazy and haze-free images. 

Currently, NiSID research is divided into two types, namely non-deep learning-based NiSID and deep learning-based NiSID. On the one hand, the prior hypotheses and statistical laws are explored \cite{zhang2017fast,zhang2020nighttime}. The maximum reflectance prior to estimate the varying ambient illumination is proposed by \cite{zhang2017fast}. The illumination estimation, color correction and image prior are integrated by \cite{zhang2014nighttime}. On the other hand, the deep learning-based architectures are designed for the NiSID task \cite{liu2023nighthazeformer,jin2023enhancing}. Liu et al. \cite{liu2023nighthazeformer} combine the dark channel and bright channel prior with the Transformer mechanism \cite{liu2021swin} into an end-to-end training flow. The gradient-adaptive convolution and glow pair synthesis are designed by Jin et al. \cite{jin2023enhancing}. Existing learning-based algorithms have achieved remarkable performance on synthetic datasets. However, these methods still lack consideration of the characteristics of nighttime hazy images.

During the day, the main source of imaging light is sunlight \cite{gui2023a}. The formation of the daytime hazy image can be described by the atmospheric scattering model \cite{gui2023a} as
\begin{equation}
  \label{eq:daytime_imaging}
  I(a) = J(a)t(a) + A(a)(1 - t(a)),
\end{equation}
where $I(a)$, $J(a)$, $t(a)$ and $A(a)$ denote the hazy image, clear image, transmission map and global atmospheric light, respectively. The $a$ means the pixel location. Meanwhile, a widely used physical model \cite{ju2021ide,koo2020nighttime} in the NiSID task is
\begin{equation}
\label{eq:nighttime_imaging}
I(a) = J(a)t(a) + A(a)(1 - t(a)) + L_{s}(a) * \varkappa(a),
\end{equation}
where $L_{s}(a)$ and $\varkappa(a)$ denote the light sources and atmospheric point spread function. As shown in Eq. \ref{eq:daytime_imaging} and Eq. \ref{eq:nighttime_imaging}, the main distinction between daytime and nighttime haze imaging is light sources \cite{li2015nighttime,wang2022variational,yang2018superpixel,liu2022multi,liu2022nighttime,ancuti2020day,dai2022flare7k}, which we consider to be the main source of the difficulty. Specifically, two outstanding issues are considered as follows.
\begin{itemize}
  \item \textbf{Localized, Coupled and Frequency Inconsistent}: As shown in Figure \ref{fig:visual_results_of_histogram_hazy_and_dehazed}, multiple active light sources may exist simultaneously. Therefore, the distortion of nighttime images, namely the haze that is mainly generated by suspended particles and liquid water droplets, the glow that is mainly produced by active light sources and the noise that is mainly caused by low intensity, is usually \textit{localized}. Meanwhile, these types of distortions are mixed throughout the image, which is \textit{coupled}. Furthermore, the haze and glow will cause the loss of high-frequency signals, while the noise belongs to high-frequency disturbance signals \cite{li2023embedding} that needs to be eliminated. This means that these distortions have \textit{inconsistent frequency characteristics}. In a word, a challenging issue is \textit{how to simultaneously handle distortions with localized, coupled and frequency inconsistent characteristics.}
  \item \textbf{Unrealistic Brightness Intensity}: Nighttime hazy datasets based on real-world images synthesized by imaging model (IM) are difficult to simulate multiple active light sources, while nighttime hazy datasets based on game engine (GE) cannot perfectly reproduce the harmonious brightness of real-world nighttime scenes. As we observed in Figure \ref{fig:visual_results_of_histogram_hazy_and_dehazed}, the dehazed images obtained under IM datasets still suffer from the glow and haze that caused by multiple light sources, but the overall brightness is realistic. The dehazed images obtained under GE dataset show less haze and glow, but the scene brightness is unrealistic. In a word, an unsolved problem faced by data-driven algorithms is \textit{how to suppress haze and glow while achieving realistic brightness.}
\end{itemize}

Therefore, we propose a semi-supervised dehazing framework that can be used for the real-world NiSID task. Firstly, the local attention \cite{liu2021swin} is adopted to learn the inductive bias in the spatial domain to suppress local distortions. A frequency spectrum dynamic filtering strategy is designed to handle distortions with inconsistent frequency characteristics. Considering the coupled of these distortions, the spatial and frequency information are integrated as a bidomain interaction module for feature extraction and image reconstruction. Secondly, aiming at suppressing distortions while achieving realistic brightness. The simulation data provided by the game engine is utilized to generate pseudo labels that can suppress haze and glow for retraining process. Then, real-world hazy images are adopted as brightness-realistic signals for the realistic brightness constraint. Overall, the main contributions of this paper are as follows.
\begin{itemize}
  \item We propose a \underline{s}patial and \underline{f}requency domain aware \underline{s}emi-supervised \underline{ni}ghttime \underline{d}ehazing network (SFSNiD). SFSNiD can remove nighttime haze that is accompanied by glow and noise. The experimental results on synthetic and real-world datasets show that the proposed method can achieve impressive performance.
  \item We design a \underline{s}patial and \underline{f}requency domain \underline{i}nformation \underline{i}nteraction (SFII) module to \textit{simultaneously handle the haze, glow and noise with localized, coupled and frequency inconsistent characteristics}. The multi-channel amplitude and phase spectrums are dynamically filtered and aggregated. The spatial and frequency domain features are integrated by local attention. 
  \item A retraining strategy and a local window-based brightness loss for semi-supervised training process are designed to \textit{suppress haze and glow while achieving realistic brightness}. The retraining strategy is based on pseudo labels. The hazy image is divided into non-overlapping windows for the calculation of local brightness map to provide realistic brightness supervision. 
\end{itemize}

\begin{figure*}
  \centering
  \includegraphics[width=17.39cm,height=6cm]{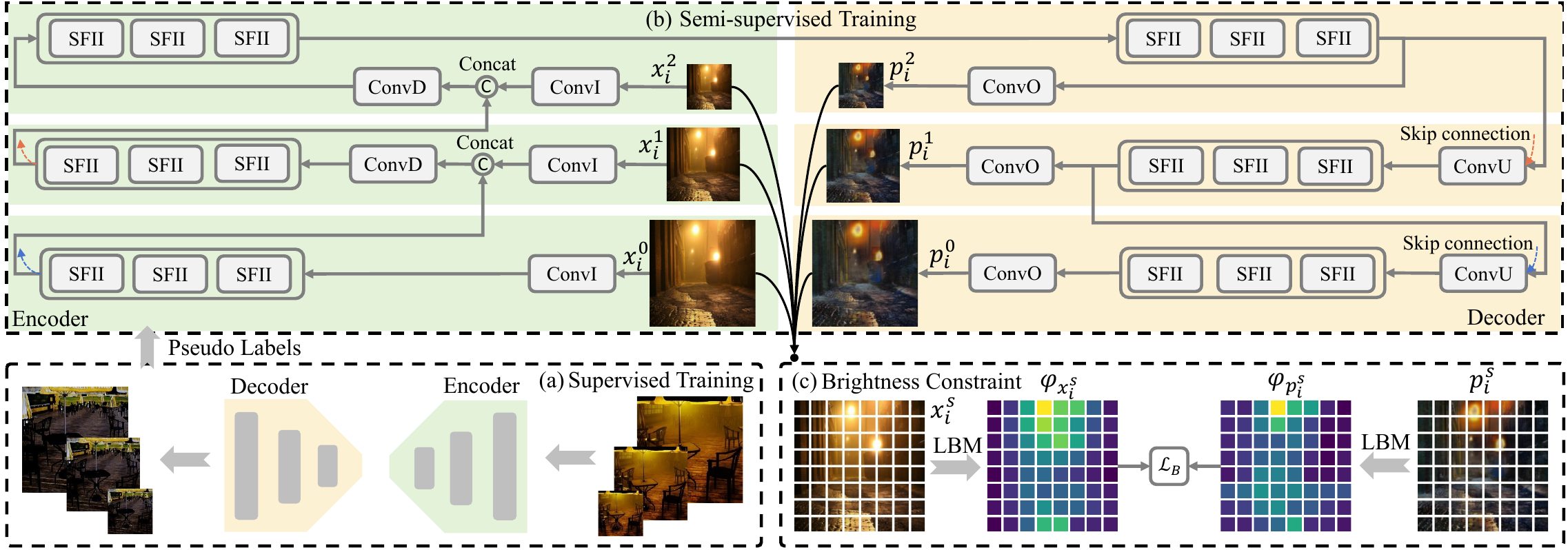}
  \caption{The overall pipeline of the proposed SFSNiD.}
  
  \label{fig:overall_network}
\end{figure*}

\section{Related Work}
\label{sec:related_work}


\subsection{Daytime Dehazing}
\label{subsec:daytime_dehazing}
A variety of effective dehazing algorithms for DaSID have been proposed. An ultra-high resolution dehazing method based on bilateral gird is proposed by 4KDehazing \cite{zheng2021ultra}. AECRNet \cite{wu2021contrastive} introduces the contrastive learning to the dehazing process. The prior information and visual attention mechanism are utilized in DeHamer \cite{guo2022image}. DF \cite{song2023vision} designs an encoder-decoder architecture which totally based on multi-head self-attention \cite{liu2021swin}. MITNet \cite{shen2023mutual} combines the mutual information-driven constraint and adaptive triple interaction strategy into a supervised training process. Although these DaSID algorithms have achieve impressive performance, they are not designed for the characteristics of nighttime hazy images, which may cause them to have certain limitations on the NiSID task \cite{liu2023nighthazeformer}.


\subsection{Nighttime Dehazing}
\label{subsec:nighttime_dehazing}
Compared with DaSID, NiSID has received fewer attention. On the one hand, the prior hypotheses and statistical laws are utilized in the non-deep learning-based NiSID methods \cite{zhang2017fast,zhang2020nighttime}. A maximum reflectance prior is proposed by MRP \cite{zhang2017fast}, which providing a way to estimate the varying ambient illumination. An optimal-scale fusion-based method is designed by OSFD \cite{zhang2020nighttime}, which utilizes a parameter estimation dehazing flow. On the other hand, the data-driven strategies \cite{li2021proposal,wang2023eulermormer} are adopted in the deep learning-based NiSID methods \cite{liu2023nighthazeformer,jin2023enhancing,yan2020nighttime}. NightHazeFormer \cite{liu2023nighthazeformer} combines the visual transformer and prior knowledge (dark channel and bright channel) into an end-to-end enhancement process. GAC \cite{jin2023enhancing} utilizes the angular point spread function to reduce the glow effect in nighttime scenes. Yan et al. \cite{yan2020nighttime} propose a strategy which decomposes the image into scene texture information and scene structure information. According to recent research, deep learning-based NiSID algorithms can achieve relatively better quantitative performance according to sufficient synthetic data. However, the haze, glow, and noise with localized, coupled and frequency inconsistent characteristics are not fully considered by these deep learning-based NiSID algorithms. 


\section{Methods}
\label{sec:methods}

The hazy domain and haze-free domain are marked as $X$ and $Y$, respectively. The synthesized hazy and haze-free image datasets are denoted $\mathcal{D}_{X}$ and $\mathcal{D}_{Y}$, which contain $N$ images, respectively. The real-world hazy image and haze-free datasets are denoted as $\mathcal{R}_{X}$ and $\mathcal{R}_{Y}$, which include $M$ images, respectively. The convolution operation is denoted as $C_{t}^{k}(\cdot)$, where the superscript $k$ and subscript $t$ denote the kernel size and stride, respectively. The $\varpi(\cdot)$, $\sigma(\cdot)$, $\delta(\cdot)$ and $sf(\cdot)$ denote the global average pooling, LeakyReLU, sigmoid and softmax operations, respectively. The input hazy images and predicted dehazed images at three scales are marked $x_{i}^{s} \in \mathcal{D}_{X}$ and $p_{i}^{s}$ respectively, where $s \in \{0, 1, 2\}$ and $i$ denotes the $i$-th example. The size of $x_{i}^{0}$, $x_{i}^{1}$ and $x_{i}^{2}$ are $H \times W \times C$, $\frac{H}{2} \times \frac{W}{2} \times C$ and $\frac{H}{4} \times \frac{W}{4} \times C$, respectively. The $H$, $W$ and $C$ denote the height, width and number of channels, respectively. The size of $p_{i}^{s}$ remains the same as $x_{i}^{s}$. The network at scale $s$ is denoted as $\Psi^{s}(\cdot)$.

\subsection{Network Structure}
The multi-scale structure \cite{cui2022selective} of the SFSNiD is shown in Figure \ref{fig:overall_network}. Two kinds of modules are included in the proposed network, namely (i) \underline{s}patial and \underline{f}requency \underline{i}nformation \underline{i}nteraction (SFII) model, (ii) \underline{c}onvolution \underline{i}nput (ConvI), \underline{c}onvolution \underline{o}utput (ConvO), \underline{c}onvolution \underline{d}ownsampling (ConvD), and \underline{c}onvolution \underline{u}psampling (ConvU). The ConvI projects the image into the feature space, while ConvO does the opposite. ConvD reduces the length and width of the feature map by half, while ConvU does the opposite. 

\begin{figure}
  \centering
  \includegraphics[width=8.2cm,height=4.5cm]{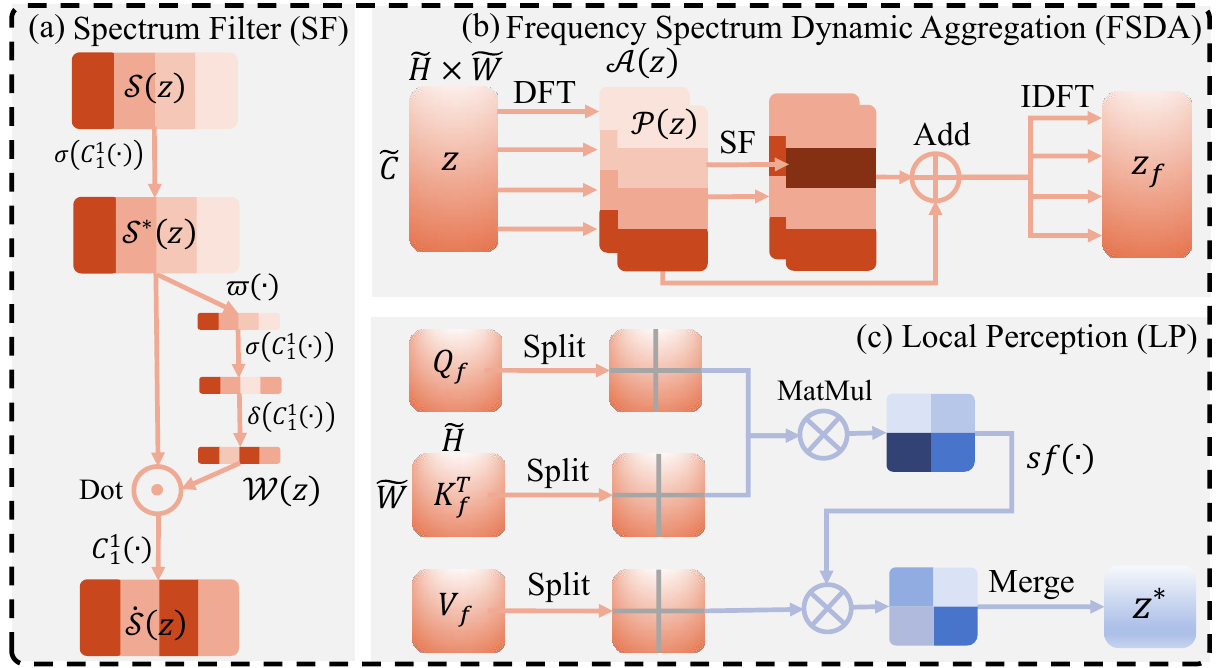}
  \caption{The sub-modules of the proposed SFII.}
  
  \label{fig:sub_blocks}
\end{figure}

\begin{figure}
  \centering
  \includegraphics[width=8.2cm,height=3.5cm]{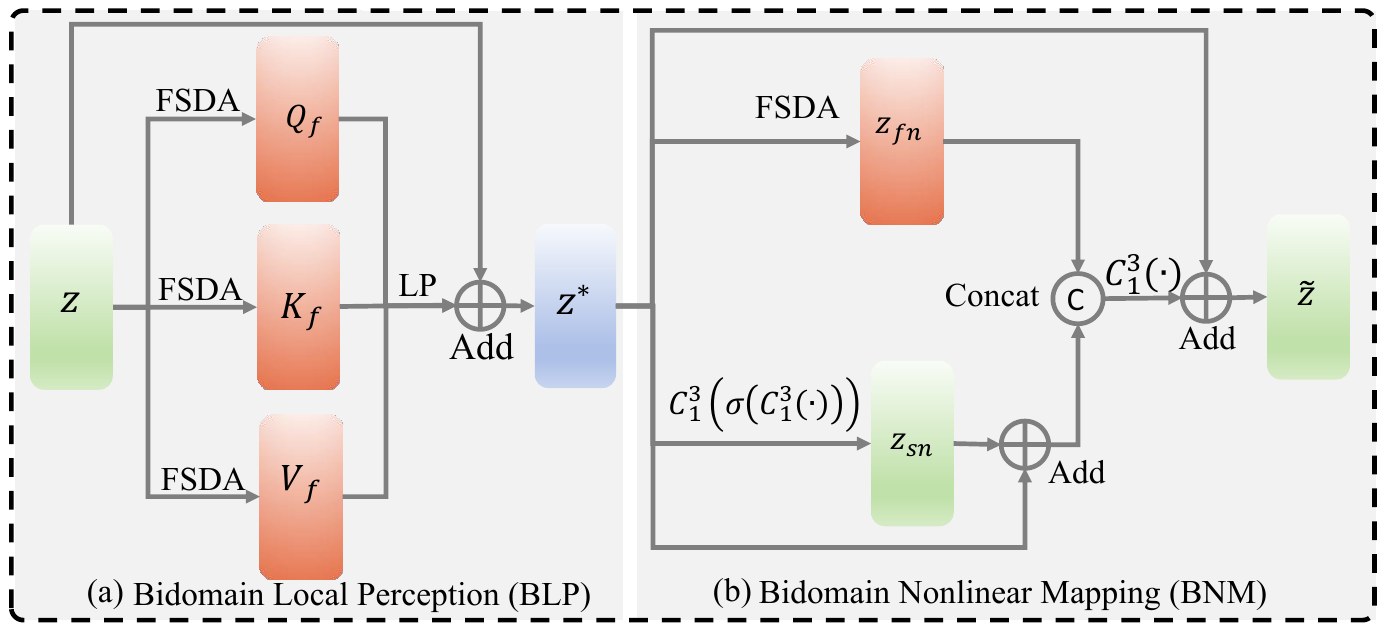}
  \caption{The overall architecture of the proposed SFII.}
  
  \label{fig:overall_block}
\end{figure}


\subsection{Spatial and Frequency Information Interaction}

\noindent
\textbf{Preliminary.} For a feature map $z \in {\mathbb{R}}^{\widetilde{H} \times \widetilde{W} \times \widetilde{C}}$, where $\widetilde{H}$, $\widetilde{W}$ and $\widetilde{C}$ denote the height, width and number of channels, respectively. We first project each of its channel $z_{\widetilde{c}}$ to the frequency domain by the Fourier \cite{guo2022exploring} transformation $\mathcal{F}$ as
\begin{equation}
  \label{eq:fft}
  \mathcal{F}(z_{\widetilde{c}})(u, v) = \sum_{h=0}^{\widetilde{H}-1} \sum_{w=0}^{\widetilde{W}-1} z_{\widetilde{c}}(h, w) e^{-j2{\pi}(\frac{h}{\widetilde{H}}u + \frac{w}{\widetilde{W}}v)},
\end{equation}
where $(h, w)$ and $(u, v)$ represent the coordinates in the spatial and frequency domain, respectively. The $\widetilde{c} \in \{0, 1, ..., \widetilde{C}\}$ denotes the channel index. Correspondingly, the $\mathcal{F}^{-1}$ is defined as the inverse Fourier transformation \cite{zhou2023fourmer}. Then, the real part $\mathcal{R}(z_{\widetilde{c}})(u, v)$ and imaginary part $\mathcal{I}(z_{\widetilde{c}})(u, v)$ can be obtained by $\mathcal{F}(z_{\widetilde{c}})(u, v)$. The amplitude spectrum $\mathcal{A}(z_{\widetilde{c}})(u, v)$ and phase spectrum $\mathcal{P}(z_{\widetilde{c}})(u, v)$ of $\mathcal{F}(z_{\widetilde{c}})(u,v)$ on the single channel can be obtained by 
\begin{equation}
  \label{eq:calculating_amplitude}
  \mathcal{A}(z_{\widetilde{c}})(u, v) = \sqrt{\mathcal{R}^{2}(z_{\widetilde{c}})(u, v) + \mathcal{I}^{2}(z_{\widetilde{c}})(u, v)},
\end{equation}
\begin{equation}
  \label{eq:calculating_phase}
  \mathcal{P}(z_{\widetilde{c}})(u, v) = \arctan [\frac{\mathcal{I}(z_{\widetilde{c}})(u, v)}{\mathcal{R}(z_{\widetilde{c}})(u, v)}].
\end{equation}

The full channel amplitude spectrum $\mathcal{A}(z)(u, v) \in {\mathbb{R}}^{\widetilde{H} \times \widetilde{W} \times \widetilde{C}}$ and phase spectrum $\mathcal{P}(z)(u, v) \in {\mathbb{R}}^{\widetilde{H} \times \widetilde{W} \times \widetilde{C}}$ can be obtained by applying the Eq. \ref{eq:fft}, Eq. \ref{eq:calculating_amplitude} and Eq. \ref{eq:calculating_phase} on each channel of $z$.

\noindent
\textbf{Frequency Spectrum Dynamic Aggregation (FSDA).}
The haze, glow and noise with inconsistent frequency characteristics can be processed in the frequency domain by dynamic spectrum filter. The amplitude spectrum and phase spectrum of different channels are aggregated by the point-wise convolution as
\begin{equation}
  \mathcal{S}^{*}(z)(u, v) = \sigma(C_{1}^{1}(\mathcal{S}(z)(u, v))), 
\end{equation}
where $S(z)(u, v) \in \{\mathcal{A}(z)(u, v), \mathcal{P}(z)(u, v)\}$. To perform channel aggregation of spectral information, the channel weight \cite{hu2018squeeze} map $\mathcal{W}$ are calculated as
\begin{equation}
  \mathcal{W}(z)(u, v) = \delta(C_{1}^{1}(\sigma(C_{1}^{1}(\varpi(S^{*}(z)(u, v)))))),
\end{equation}
where $\mathcal{W}(z)(u, v) \in {\mathbb{R}}^{1 \times 1 \times \widetilde{C}}$. Then the channel weight map is applied to the frequency spectrum as 
\begin{equation}
  \dot{\mathcal{S}}(z)(u, v) = C_{1}^{1}(\mathcal{W}(z)(u, v) \cdot \mathcal{S}^{*}(z)(u, v)),
\end{equation}
where the \underline{s}pectrum \underline{f}ilter (SF) of $\dot{S}(z)(u, v)$ is shown in Figure \ref{fig:sub_blocks}-(\textcolor{red}{a}). The filtering operation is performed by the residual connection, the filtered component is obtained by
\begin{equation}
  \widetilde{\mathcal{S}}(z)(u, v) = \dot{\mathcal{S}}(z)(u, v) + \mathcal{S}(z)(u, v).
\end{equation}

The filtered $\widetilde{\mathcal{A}}(z)(u, v)$ and $\widetilde{\mathcal{P}}(z)(u, v)$ can be obtained based on the processing flow from $\mathcal{S}(z)(u, v)$ to $\widetilde{S}(z)(u, v)$. Then, the real and imaginary parts are obtained by
\begin{equation}
  \widetilde{\mathcal{R}}(z)(u, v) = \widetilde{\mathcal{A}}(z)(u, v) \cdot \cos{\widetilde{\mathcal{P}}(z)(u, v)},
\end{equation}
\begin{equation}
  \widetilde{\mathcal{I}}(z)(u, v) = \widetilde{\mathcal{A}}(z)(u, v) \cdot \sin{\widetilde{\mathcal{P}}(z)(u, v)}.
\end{equation}

After dynamic parameter learning in the frequency domain, we remap the feature map to the spatial domain as
\begin{equation}
  \label{eq:ifft}
  z_{f} = \mathcal{F}^{-1}(\widetilde{\mathcal{R}}(z)(u, v), \widetilde{\mathcal{I}}(z)(u, v)),
\end{equation}
where $z_{f} \in {\mathbb{R}}^{\widetilde{H} \times \widetilde{W} \times \widetilde{C}}$. The Fourier transformation and inverse Fourier transformation can be implemented using DFT and IDFT algorithms \cite{frigo1998fftw,hou2023bidomain,zhou2023pan}. Here, we define the calculation from Eq. \ref{eq:fft} to Eq. \ref{eq:ifft} as \underline{f}requency \underline{s}pectrum \underline{d}ynamic \underline{a}ggregation (FSDA), which represent the processing flow from $z$ to $z_{f}$ that is shown in Figure \ref{fig:sub_blocks}-(\textcolor{red}{b}). For convenience, the FSDA is denoted as $\mathcal{FS}(\cdot)$.

\noindent
\textbf{Frequency Domain Projection (FDP).} To deal with distortions in the frequency domain, we first introduce frequency domain interactions before computing local inductive bias. For the input feature map $z \in \mathbb{R}^{\widetilde{H} \times \widetilde{W} \times \widetilde{C}}$, it is processed by the layer normalization operation ($LN(\cdot)$) \cite{liu2021swin} to obtain the normalized feature $z_{l} = LN(z)$. Then, the normalized feature $z_{l}$ is projected into $Q_f$ (query), $K_f$ (key) and $V_f$ (value) by the projection in the frequency domain as
\begin{equation}
  Q_{f} = \mathcal{FS}_{Q}(z_{l}), K_{f} = \mathcal{FS}_{K}(z_{l}), V_{f} = \mathcal{FS}_{V}(z_{l}),
\end{equation}
where the $\mathcal{FS}_{Q}(\cdot)$, $\mathcal{FS}_{K}(\cdot)$ and $\mathcal{FS}_{V}(\cdot)$ denote three independent projection operation with learnable parameters, respectively. The generation process of the $Q_{f}$, $K_{f}$ and $V_{f}$ is denoted as the \underline{f}requency \underline{d}omain \underline{p}rojection (FDP).

\noindent
\textbf{Bidomain Local Perception (BLP).} After obtaining the features $Q_{f}$, $K_{f}$ and $V_{f}$ which consider the information in frequency domain, we perform spatial domain learning on the features from a local perspective. The self-attention \cite{liu2021swin} with \underline{l}ocal \underline{p}erception (LP) that is shown in Figure \ref{fig:sub_blocks}-(\textcolor{red}{c}) is computed within $8 \times 8$ non-overlapping windows as 
\begin{equation}
  \label{eq:attention}
  \mathcal{AT}(Q_{f}, K_{f}, V_{f}) = sf(\frac{Q_{f} \otimes K_{f}^{T}}{\sqrt{d}} + B) \otimes V_{f},
\end{equation}
where $d$ and $B$ denote the dimensionality and position bias, respectively. The $\otimes$ denotes the matrix multiplication (MatMul).  Information is transferred by the residual connection
\begin{equation}
  z^{*} = \mathcal{AT}(Q_{f}, K_{f}, V_{f}) + z,
\end{equation}
where the calculation from $z$ to $z^{*}$ is marked as \underline{b}idomain \underline{l}ocal \underline{p}erception (BLP), which is shown in Figure \ref{fig:overall_block}-(\textcolor{red}{a}).

\noindent
\textbf{Bidomain Nonlinear Mapping (BNM).} The computation of window attention does not provide nonlinear representation capabilities. Therefore, we use the frequency and spatial domain interaction module to learn nonlinear mapping. The FSDA is used to provide the frequency domain information. Besides, a residual block which consists of $C_{1}^{3}(\sigma(C_{1}^{3}(\cdot)))$ is used to provide the spatial interaction. The immediate feature $z^{*}$ is fed into the frequency nonlinear mapping branch and spatial nonlinear mapping branch, as
\begin{equation}
  z_{fn} = \mathcal{FS}_{A}(z^{*}),
\end{equation}
\begin{equation}
 z_{sn} = C_{1}^{3}(\sigma(C_{1}^{3}(z^{*}))),
\end{equation}
where the subscript $A$ in $\mathcal{FS}_{A}(\cdot)$ means the frequency interaction performed after the attention operation. Then frequency domain and spatial domain features are fused as the final nonlinear mapping output by
\begin{equation}
  \widetilde{z} = C_{1}^{3}([z_{fn}, z_{sn} + z^{*}]) + z^{*},
\end{equation}
where the $[\cdot, \cdot]$ denotes the channel concatenation. The calculation from $z^{*}$ to $\widetilde{z}$ is marked as the \underline{b}idomain \underline{n}onlinear \underline{m}apping (BNM), which is shown in Figure \ref{fig:overall_block}-(\textcolor{red}{b}).

\noindent
\textbf{Spatial and Frequency Information Interaction (SFII).}
As shown in Figure \ref{fig:overall_block}, the calculation process from $z$ to $\widetilde{z}$ is called \underline{s}patial and \underline{f}requency \underline{i}nformation \underline{i}nteraction (SFII). The proposed SFII aggregates spatial domain information and frequency domain information from a local perspective.

\subsection{Spatial and Frequency Loss}
The supervised loss consists of two parts, namely the pixel-by-pixel loss in geometric space and the frequency domain loss obtained by Fourier transform \cite{cui2022selective}. By sampling $x_{i}^{s} \in \mathcal{D}_{X}$ and $y_{i}^{s} \in \mathcal{D}_{Y}$, the losses calculated at three scales are
\begin{equation}
  \label{eq:loss_supervised_L1}
  \mathcal{L}_{G} = \sum_{s=0}^{2} \lambda_{g} \cdot \sum_{i=0}^{N-1} ||\Psi^{s}(x_{i}^{s}) - y_{i}^{s}||_{1}, 
\end{equation}
\begin{equation}
  \label{eq:loss_supervised_FFT}
  \mathcal{L}_{F} = \sum_{s=0}^{2} \lambda_{f} \cdot \sum_{i=0}^{N-1} ||\mathcal{F}(\Psi^{s}(x_{i}^{s})) - \mathcal{F}(y_{i}^{s})||_1,
\end{equation}
where $\lambda_{g}$ and $\lambda_{f}$ denote weight factors. 

\subsection{Retraining and Realistic Brightness Loss}

\noindent
\textbf{Pseudo-label Fusion Retraining}. There are inherent domain discrepancy between synthetic hazy images and real-world hazy images. Therefore, we adopt a retraining strategy which utilizes pesudo labels. Pseudo labels $\mathcal{R}_{Y}^{P}$ are obtained based on the model trained on synthetic datasets. We put the original synthetic dataset $\{\mathcal{D}_{X}, \mathcal{D}_{Y}\}$ and the pseudo-labeled dataset $\{\mathcal{R}_{X}, \mathcal{R}_{Y}^{P}\}$ into the network simultaneously for retraining. Supervised losses Eq. \ref{eq:loss_supervised_L1} and Eq. \ref{eq:loss_supervised_FFT} are used in the retraining process at three scales.

\noindent
\textbf{Prior Brightness Constraint.}  We conduct a quantitative statistics on the brightness of nighttime hazy and clear images provided by \cite{jin2023enhancing}. The brightness intensity corresponding to $x_{i}^{0} \in \mathcal{R}_{X}$ and $y_{i}^{0} \in \mathcal{R}_{Y}$ are $\mu(x_{i}^{0})$ and $\mu(y_{i}^{0})$, respectively, where $\mu(\cdot)$ denote the average pixel value across three channels. We randomly select $\hat{M} = \frac{M}{2}$ images from the dataset multiple times, and we get
\begin{equation}
  \label{eq:prior_brightness_constraint}
  \sum_{i=0}^{\hat{M}-1} \mu(y_{i}^{0}) \textless \sum_{i=0}^{\hat{M}-1} \mu(x_{i}^{0}).
\end{equation}

Therefore, we assume the brightness of the dehazed image $p_{i}^{s}$ should be lower than that of the $x_{i}^{s}$. This assumption is consistent with the imaging model Eq. \ref{eq:nighttime_imaging}.

\noindent
\textbf{Local Brightness Map (LBM).} We divide the image into non-overlapping local windows. The width and height of each square window is denoted as $\gamma^{s}$, where $s \in \{0, 1, 2\}$. The value in \underline{l}ocal \underline{b}rightness \underline{m}ap (LBM) $\varphi_{x_i^s}$ that corresponding to $x_{i}^{s}$ is obtained by
\begin{equation}
  \varphi_{x_i^s}(\hat{h}, \hat{w}) = \frac{1}{3 (\gamma^{s})^{2}} \sum_{c=0}^{2} \sum_{h=\hat{h} \cdot \gamma^{s}}^{(\hat{h}+1) \cdot \gamma^{s}} \sum_{w=\hat{w} \cdot \gamma^{s}}^{(\hat{w}+1) \cdot \gamma^{s}} x_{i}^{s}(h, w, c),
\end{equation}
where ($\hat{h}, \hat{w}$) and $(h, w)$ denote the pixel index of $\varphi_{x_i^s}$ and $x_{i}^{s}$, respectively. Meanwhile, the local brightness map $\varphi_{p_{i}^{s}}$ corresponding to $p_{i}^{s}$ is defined in the same way. As shown in Figure \ref{fig:overall_network}-(\textcolor{red}{c}), the locations with high brightness may be active light sources or objects close to the light source, while the locations with low brightness may be objects and backgrounds far away from the light source.

\begin{figure*}
  \centering
  \footnotesize
  \includegraphics[width=2.4cm,height=1.5cm]{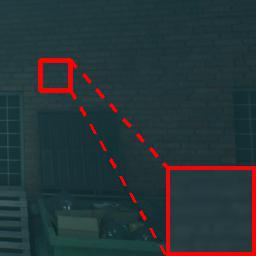}
  \includegraphics[width=2.4cm,height=1.5cm]{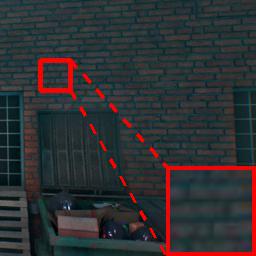}
  \includegraphics[width=2.4cm,height=1.5cm]{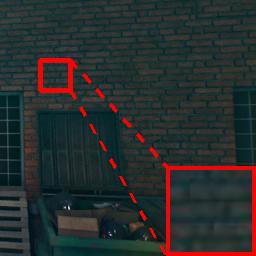}
  \includegraphics[width=2.4cm,height=1.5cm]{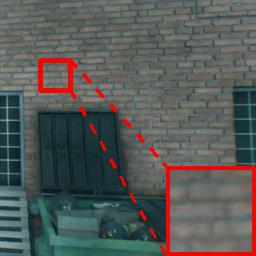}
  \includegraphics[width=2.4cm,height=1.5cm]{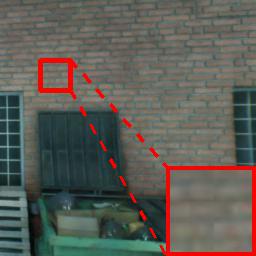}
  \includegraphics[width=2.4cm,height=1.5cm]{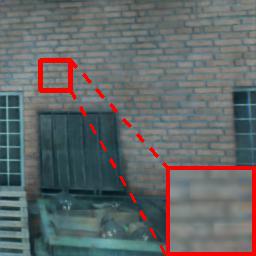}
  \includegraphics[width=2.4cm,height=1.5cm]{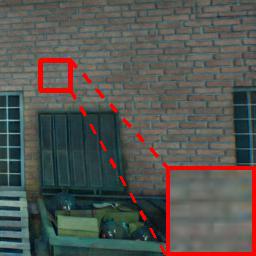}
  \leftline{\hspace{0.8cm} (a) Hazy \hspace{1.3cm} (b) MRP \hspace{1.3cm} (c) OSFD \hspace{1.3cm} (d) GD \hspace{1.3cm} (e) MSBDN \hspace{0.8cm} (f) 4KDehazing \hspace{0.7cm} (g) AECRNet}
  \\
  \vspace{0.1cm}
  \includegraphics[width=2.4cm,height=1.5cm]{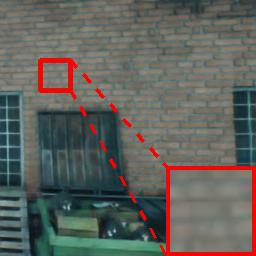}
  \includegraphics[width=2.4cm,height=1.5cm]{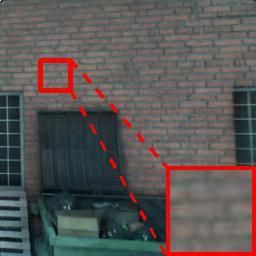}
  \includegraphics[width=2.4cm,height=1.5cm]{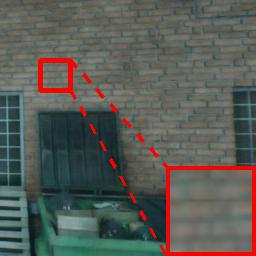}
  \includegraphics[width=2.4cm,height=1.5cm]{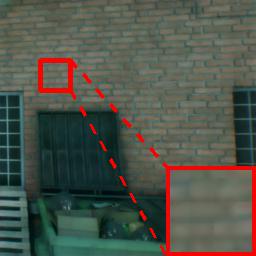}
  \includegraphics[width=2.4cm,height=1.5cm]{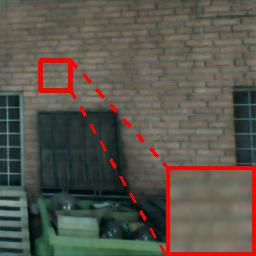}
  \includegraphics[width=2.4cm,height=1.5cm]{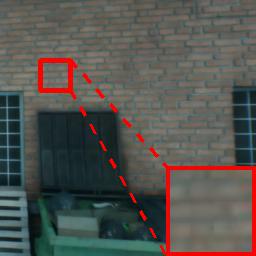}
  \includegraphics[width=2.4cm,height=1.5cm]{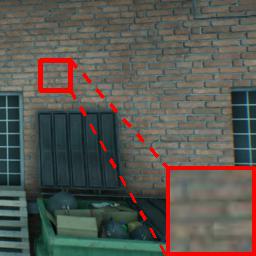}

  \leftline{\hspace{0.5cm} (h) DeHamer \hspace{0.9cm} (i) FSDGN \hspace{1.3cm} (j) DF \hspace{1.4cm} (k) MITNet \hspace{1.1cm} (l) Fourmer \hspace{1.1cm} (m) Ours \hspace{1.3cm} (n) Label}
  \caption{Visual results on synthetic dataset \cite{liu2023nighthazeformer}.}
  \label{fig:visual_results_UNREAL}
\end{figure*}

\begin{table*}[t]
  \scriptsize
  \setlength{\tabcolsep}{1.5mm}
	\renewcommand{\arraystretch}{1.2}
    \centering
    \caption{Quantitative results on datasets that generated by imaging model.}
    \label{tab:nighttime_dehazing_results_synthesized}
    
    \begin{tabular}{cc@{\hspace{0.27cm}}cc@{\hspace{0.27cm}}cc@{\hspace{0.27cm}}cc@{\hspace{0.27cm}}cc@{\hspace{0.27cm}}cc@{\hspace{0.27cm}}cc@{\hspace{0.27cm}}c}
      \hline
        \multirow{2}{*}{Methods} & \multicolumn{2}{c}{NHR} & \multicolumn{2}{c}{NHM} & \multicolumn{2}{c}{NHCL} & \multicolumn{2}{c}{NHCM} & \multicolumn{2}{c}{NHCD} & \multicolumn{2}{c}{NightHaze} & \multicolumn{2}{c}{YellowHaze}\\
      \cline{2-15}
       & SSIM$\uparrow$ & PSNR$\uparrow$ & SSIM$\uparrow$ & PSNR$\uparrow$ & SSIM$\uparrow$ & PSNR$\uparrow$ & SSIM$\uparrow$ & PSNR$\uparrow$ & SSIM$\uparrow$ & PSNR$\uparrow$  & SSIM$\uparrow$ & PSNR$\uparrow$ & SSIM$\uparrow$ & PSNR$\uparrow$ \\
  
      \hline   
      
      MRP (CVPR 2017)         & 0.776 & 19.848 & 0.666 & 15.993 & 0.747 & 22.497 & 0.693 & 20.494 & 0.624 & 17.651 & 0.295 & 12.138 & 0.249 & 13.473 \\
      GD (ICCV 2019)          & 0.969 & 30.107 & 0.861 & 20.689 & 0.973 & 36.506 & 0.958 & 34.448 & 0.932 & 31.509 & 0.832 & 25.324 & 0.915 & 27.410 \\
      OSFD (ACMMM 2020)        & 0.808 & 21.028 & 0.722 & 18.491 & 0.786 & 22.329 & 0.739 & 20.929 & 0.672 & 18.501 & 0.304 & 13.387 & 0.259 & 14.775 \\
      MSBDN (CVPR2020)       & 0.970 & 31.335 & 0.818 & 20.514 & 0.965 & 35.963 & 0.938 & 32.848 & 0.903 & 30.475 & 0.950 & 33.156 & 0.921 & 29.834 \\
      4KDehazing (CVPR2021)   & 0.950 & 28.613 & 0.830 & 20.429 & 0.967 & 35.006 & 0.958 & 35.162 & 0.912 & 30.048 & 0.850 & 26.562 & 0.861 & 25.835 \\
      AECRNet (CVPR 2021)     & 0.915 & 24.864 & 0.817 & 19.420 & 0.951 & 33.183 & 0.943 & 33.498 & 0.890 & 28.742 & 0.946 & 32.344 & 0.937 & 29.417 \\
      DeHamer (CVPR 2022)     & 0.966 & 31.017 & 0.823 & 23.095 & 0.966 & 36.038 & 0.944 & 33.908 & 0.915 & 31.389 & 0.954 & 33.432 & 0.931 & 30.334 \\
      FSDGN  (ECCV 2022)      & 0.975 & 32.072 & 0.874 & 21.415 & 0.972 & 36.432 & 0.952 & 33.723 & 0.922 & 31.559 & 0.948 & 33.521 & 0.955 & \textbf{33.062} \\
      DF (TIP 2023) & 0.969 & 31.644 & 0.896 & 23.207 & 0.975 & 37.383 & 0.960 & 35.038 & 0.934 & 32.079 & 0.931 & 31.489 & 0.948 & 32.244 \\
      MITNet (ACMMM 2023)      & 0.974 & 31.969 & 0.859 & 20.884 & 0.969 & 35.794 & 0.945 & 32.849 & 0.916 & 30.628 & 0.946 & 34.114 & 0.932 & 31.186 \\    
      Fourmer (ICML 2023)     & 0.969 & 31.660 & 0.862 & 21.423 & 0.963 & 35.714 & 0.943 & 33.201 & 0.928 & 32.103 & 0.949 & 33.419 & 0.958 & 31.978 \\
      Ours         & \textbf{0.978} & \textbf{33.180} & \textbf{0.905} & \textbf{23.705} & \textbf{0.979} & \textbf{38.146} & \textbf{0.968} & \textbf{36.146} & \textbf{0.951} & \textbf{34.001} & \textbf{0.968} & \textbf{35.527} & \textbf{0.965} & 32.981 \\
  
    \hline
    \end{tabular}
  \end{table*}

\noindent 
\textbf{Realistic Brightness Loss.} The brightness of hazy images is approximately globally realistic, so it can be used to supervise the brightness of dehazed images. As we observed in Eq. \ref{eq:prior_brightness_constraint}, the brightness of the dehazed image should be lower than that of the hazy image. Meanwhile, in order to ensure the relative numerical relationship between areas with high brightness and low brightness before and after dehazing, we use a power function with monotonically increasing properties to process the $\varphi_{x_i^s}(\hat{h}, \hat{w})$, as
\begin{equation}
  \widetilde{\varphi}_{x_i^s}(\hat{h}, \hat{w}) = (\varphi_{x_i^s}(\hat{h}, \hat{w}))^{\kappa},
\end{equation}
where $\kappa \geq 1$ is the brightness intensity coefficient. The realistic brightness constraint within one single window is 
\begin{equation}
  \mathcal{L}_{B}^{p_{i}^{s}}(\hat{h}, \hat{w}) = (\varphi_{p_i^s}(\hat{h}, \hat{w}) - \xi \cdot \widetilde{\varphi}_{x_i^s}(\hat{h}, \hat{w}))^{2},
\end{equation}
where $\xi$ is a hyperparameter. The realistic brightness loss calculated over all windows is 
\begin{equation}
  \mathcal{L}_{B} = \sum_{s=0}^{2} \frac{\lambda_{b}}{\hat{N} \hat{W}^{s} \hat{H}^{s}} \cdot \sum_{i=0}^{\hat{N}-1} \sum_{\hat{h}=0}^{\hat{H}^{s}-1} \sum_{\hat{w}=0}^{\hat{W}^{s}-1} \mathcal{L}_{B}^{p_{i}^{s}}(\hat{h}, \hat{w}),
\end{equation}
where $\hat{W}^{s} = W^{s} / \gamma^{s}$, $\hat{H}^{s} = H^{s} / \gamma^{s}$. And $\hat{N} = N + M$. The $\lambda_{b}$ denotes the weights of scale loss of $\mathcal{L}_{B}$.

\subsection{Total Loss}
The overall loss is a combination of supervised and semi-supervised losses, which is
\begin{equation}
  \mathcal{L} = \mathcal{L}_{G} + \alpha \mathcal{L}_{F} + \beta \mathcal{L}_{B},
\end{equation}
where $\alpha$ and $\beta$ are the weights of the frequency domain loss and the realistic brightness loss, respectively.

\section{Experiments}
\label{experiments}

\subsection{Experiment Setting}

\noindent
\textbf{Datasets}. To comprehensively compare the performance of different algorithms, we conducted experiments on both synthetic and real-world datasets. The synthetic datasets include GTA5 \cite{yan2020nighttime}, UNREAL-NH \cite{liu2023nighthazeformer}, \{NHR, NHM, HNCL, NHCM, NHCD\} \cite{zhang2020nighttime} and \{NightHaze, YellowHaze\} \cite{liao2018hdp}. The real-world nighttime haze (RWNH) is provided by \cite{jin2023enhancing}. Since the brightness level of the ground-truth label in the UNREAL-NH is close to daytime, we adjust the brightness of the hazy image and corresponding label to the level of the nighttime low-light image by the Gamma correction \cite{ren2019low} for the evaluation of the RWNH.

\noindent
\textbf{Comparison Methods and Evaluation Metrics.} MRP \cite{zhang2017fast}, GD \cite{liu2019griddehazenet}, OSFD \cite{zhang2020nighttime}, MSBDN \cite{dong2020multi}. 4KDehazing \cite{zheng2021ultra}, AECRNet \cite{wu2021contrastive}, DeHamer  \cite{guo2022image}, FSDGN \cite{yu2022frequency}, DF \cite{song2023vision}, MITNet \cite{shen2023mutual} and Fourmer \cite{zhou2023fourmer} are used as comparisons. PSNR \cite{shen2023adaptive,li2023embedding,shao2020domain} and SSIM \cite{wang2004image,guo2020zero} are used to evaluate the performance on labeled datasets. BRISQUE \cite{wu2023ridcp} and MUSIQ \cite{huang2023contrastive,ke2021musiq} are computed to evaluate the performance on unlabeled dataset. The $\uparrow$ represents a larger value, a higher quality, while $\downarrow$ represents a larger value, a lower quality.

\noindent
\textbf{Implementation Details.} The batch size is chosen as $4$. The image size is set to $256 \times 256 \times 3$. The learning rate is initialized to 0.0001 and linearly decays by a factor of 0.95 every 10 epochs. The Adam ($\beta_{1}=0.9$, $\beta_{2}=0.999$) is used. The $\lambda_{g}$, $\lambda_{f}$ and $\lambda_{b}$ are all set to $1$. The $\alpha$ and $\beta$ are set to $0.1$ and $20$, respectively. The window size $\gamma^{s}$ are set to 16, 8 and 4, where $s \in \{0, 1, 2\}$, respectively. The coefficient $\xi$ and $\kappa$ is set to $1$ and $1.3$, respectively. The proposed model is implemented by PyTorch and trained on the single NVIDIA RTX 4090 platform.

\begin{figure*}
  \centering
  \footnotesize
  \includegraphics[width=2.8cm,height=1.5cm]{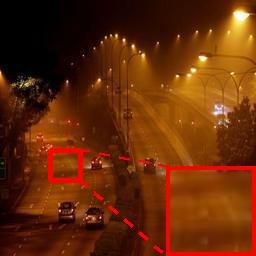}
  \includegraphics[width=2.8cm,height=1.5cm]{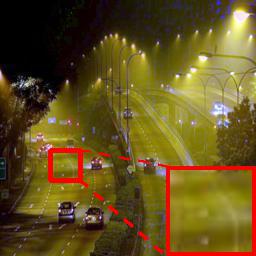}
  \includegraphics[width=2.8cm,height=1.5cm]{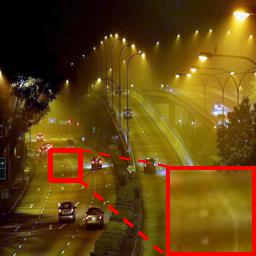}
  \includegraphics[width=2.8cm,height=1.5cm]{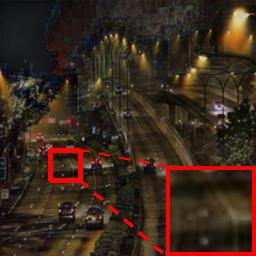}
  \includegraphics[width=2.8cm,height=1.5cm]{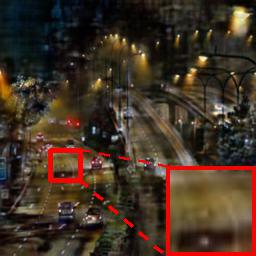}
  \includegraphics[width=2.8cm,height=1.5cm]{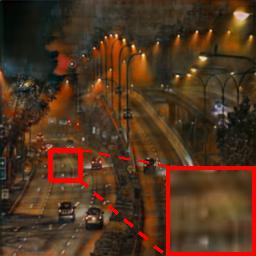}
  \\
  \includegraphics[width=2.8cm,height=1.5cm]{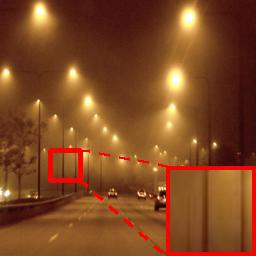}
  \includegraphics[width=2.8cm,height=1.5cm]{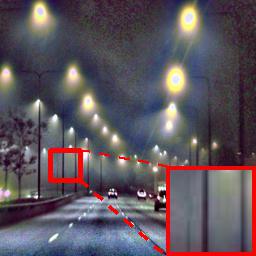}
  \includegraphics[width=2.8cm,height=1.5cm]{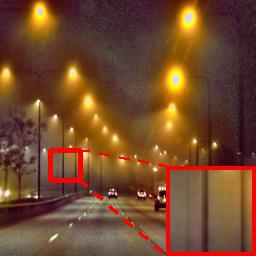}
  \includegraphics[width=2.8cm,height=1.5cm]{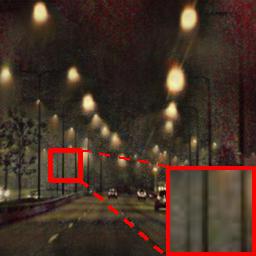}
  \includegraphics[width=2.8cm,height=1.5cm]{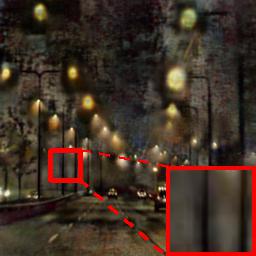}
  \includegraphics[width=2.8cm,height=1.5cm]{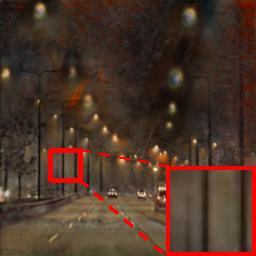}
  
  \leftline{\hspace{1.0cm} (a) Hazy \hspace{1.8cm} (b) MRP \hspace{1.6cm} (c) OSFD \hspace{1.8cm} (d) GD \hspace{1.6cm} (e) MSBDN \hspace{1.2cm} (f) 4KDehazing \hspace{0.2cm}}
  
  \vspace{0.1cm}
  \includegraphics[width=2.8cm,height=1.5cm]{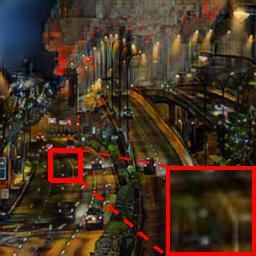}
  \includegraphics[width=2.8cm,height=1.5cm]{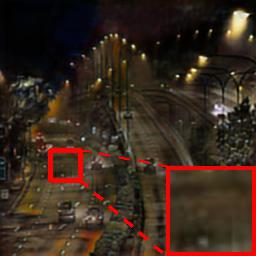}
  \includegraphics[width=2.8cm,height=1.5cm]{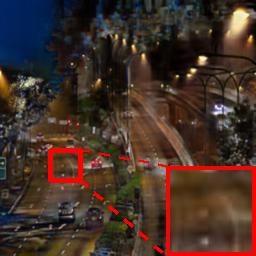}
  \includegraphics[width=2.8cm,height=1.5cm]{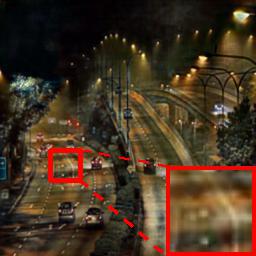}
  \includegraphics[width=2.8cm,height=1.5cm]{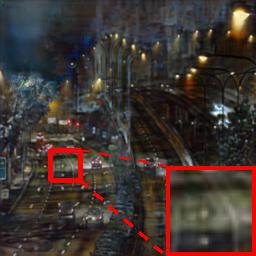}
  \includegraphics[width=2.8cm,height=1.5cm]{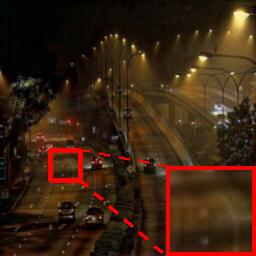}
  \\
  \includegraphics[width=2.8cm,height=1.5cm]{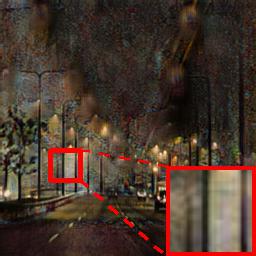}
  \includegraphics[width=2.8cm,height=1.5cm]{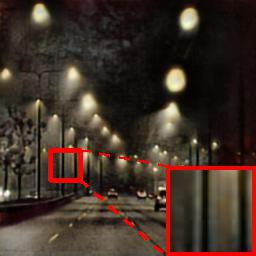}
  \includegraphics[width=2.8cm,height=1.5cm]{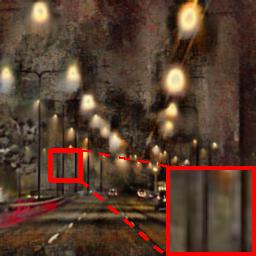}
  \includegraphics[width=2.8cm,height=1.5cm]{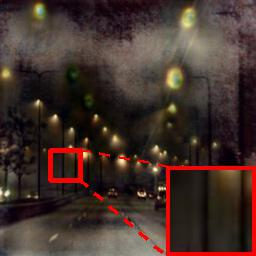}
  \includegraphics[width=2.8cm,height=1.5cm]{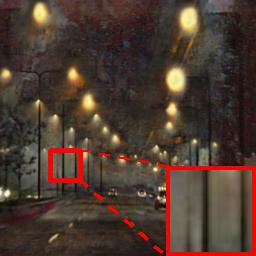}
  \includegraphics[width=2.8cm,height=1.5cm]{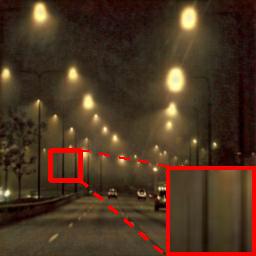}

  \leftline{\hspace{0.6cm} (g) AECRNet \hspace{1.2cm} (h) DeHamer \hspace{1.8cm} (i) DF \hspace{1.7cm} (j) MITNet \hspace{1.5cm} (k) Fourmer \hspace{1.5cm} (l) Ours}
  \caption{Visual results on real-world hazy images \cite{jin2023enhancing}.}
  \label{fig:visual_results_Real}
\end{figure*}

\begin{figure*}
  \centering
  \footnotesize
  \includegraphics[width=1.67cm,height=1.3cm]{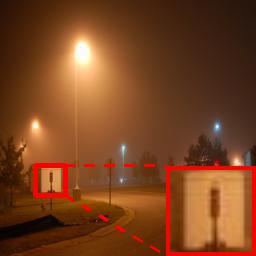}
  \includegraphics[width=1.67cm,height=1.3cm]{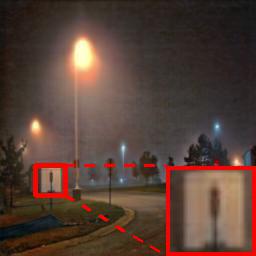}
  \includegraphics[width=1.67cm,height=1.3cm]{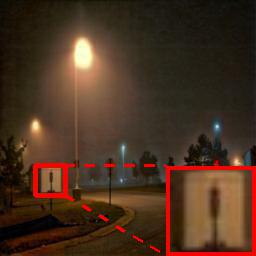}
  \includegraphics[width=1.67cm,height=1.3cm]{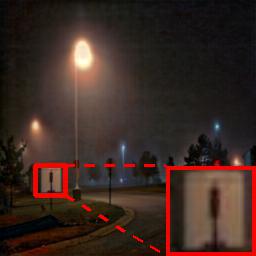}
  \includegraphics[width=1.67cm,height=1.3cm]{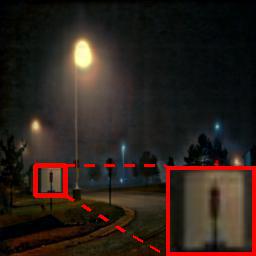}
  \includegraphics[width=1.67cm,height=1.3cm]{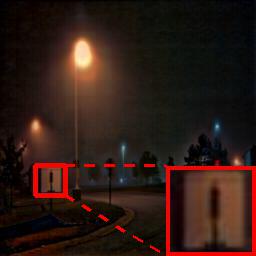}
  \includegraphics[width=1.67cm,height=1.3cm]{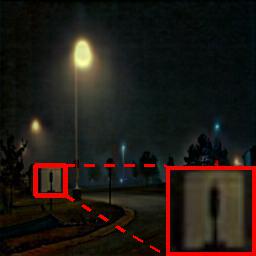}
  \includegraphics[width=1.67cm,height=1.3cm]{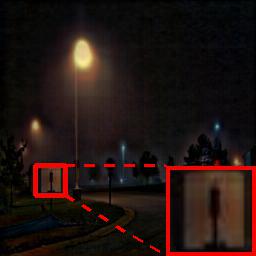}
  \includegraphics[width=1.67cm,height=1.3cm]{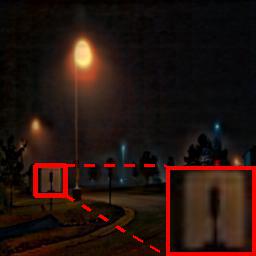}
  \includegraphics[width=1.67cm,height=1.3cm]{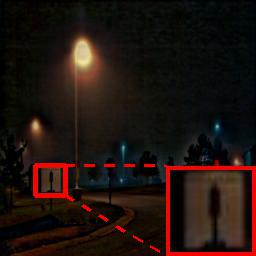}
  \\
  \includegraphics[width=1.67cm,height=1.3cm]{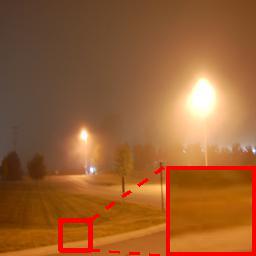}
  \includegraphics[width=1.67cm,height=1.3cm]{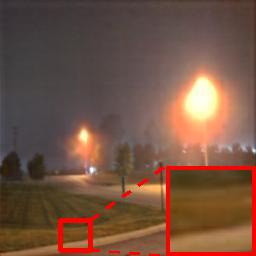}
  \includegraphics[width=1.67cm,height=1.3cm]{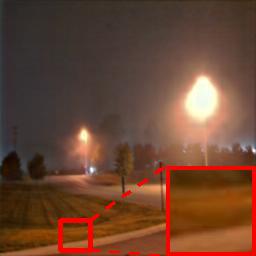}
  \includegraphics[width=1.67cm,height=1.3cm]{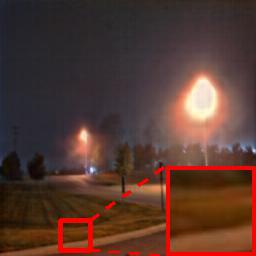}
  \includegraphics[width=1.67cm,height=1.3cm]{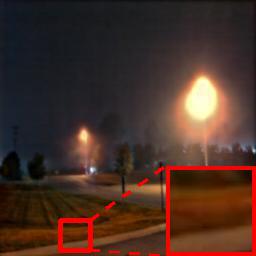}
  \includegraphics[width=1.67cm,height=1.3cm]{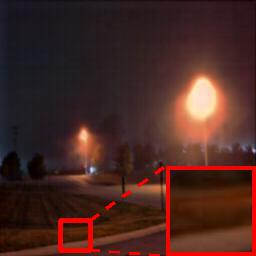}
  \includegraphics[width=1.67cm,height=1.3cm]{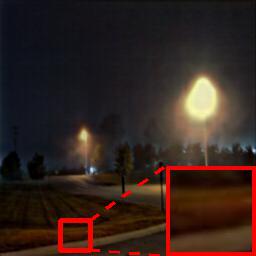}
  \includegraphics[width=1.67cm,height=1.3cm]{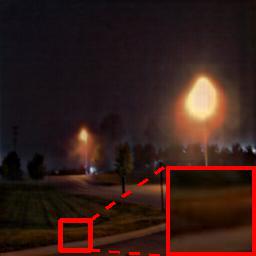}
  \includegraphics[width=1.67cm,height=1.3cm]{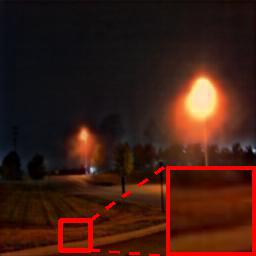}
  \includegraphics[width=1.67cm,height=1.3cm]{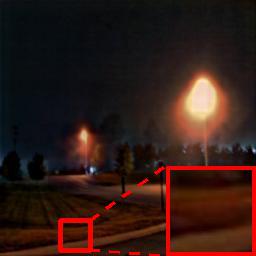}

  \leftline{\hspace{0.4cm} (a) Hazy \hspace{0.3cm} (b) $\kappa=1.0$ \hspace{0.25cm} (c) $\kappa=1.3$ \hspace{0.3cm} (d) $\kappa=1.5$ \hspace{0.3cm} (e) $\kappa=1.8$ \hspace{0.3cm} (f) $\kappa=2.0$ \hspace{0.25cm} (g) $\kappa=2.3$ \hspace{0.3cm} (h) $\kappa=2.5$ \hspace{0.25cm} (i) $\kappa=2.8$ \hspace{0.25cm} (j) $\kappa=3.0$}
  \caption{Dehazed images obtained under different $\kappa$.}
  \label{fig:visual_results_differ_brightness_intensity_coefficient_kappa}
\end{figure*}

\subsection{Comparison with State-of-the-art Algorithms}
\noindent
\textbf{Evaluation on Synthetic Datasets.} Table \ref{tab:nighttime_dehazing_results_synthesized} and Table \ref{tab:nighttime_dehazing_results_game_engine} show the quantitative dehazing results obtained by state-of-the-art methods. Figure \ref{fig:visual_results_UNREAL} shows the corresponding visual results. The quantitative and visual results demonstrate that the proposed methods achieve an overall better performance than state-of-the-art algorithms.

\noindent
\textbf{Evaluation on Real-world Datasets.} Table \ref{tab:nighttime_dehazing_results_game_engine} shows the evaluation results of real-world dehazed images. It is worth pointing out that existing research \cite{gui2023a} proposes that the reliability of no-reference metrics in the dehazing task is lower than that of full-reference metrics. Figure \ref{fig:visual_results_Real} shows that the details of the dehazed results obtained by our method are visually better. Meanwhile, the brightness of the dehazed images obtained by most comparison algorithms is obvious unrealistic, while the brightness of the dehazed images obtained by our algorithm is approximately globally realistic. 

\noindent
\textbf{Overall Evaluation.} According to the quantitative and visual results on synthetic and real-world datasets, the proposed SFSNiD achieves overall better performance. More results are placed at Supplementary Materials.


\subsection{Ablation Study and Discussions}

\noindent
\textbf{Spatial and Frequency Information Interaction.} 
The \underline{s}patial and \underline{f}requency \underline{i}nformation \underline{i}nteraction (SFII) modules and naive convolution module are used in the proposed SFSNiD. In order to prove the usefulness of the FDP, LP and BNM that contained in the SFII, ablation experiments for different sub-blocks are performed. The ablation experiment on the proposed SFII includes (i) removing the FDP, (ii) removing the LP, (iii) removing the frequency domain processing in BNM, and (iv) removing the spatial domain process in BNM. These four settings are denoted $R1$, $R2$, $R3$ and $R4$, respectively. Table \ref{tab:ablation_on_basic_block} shows the ablation results under different settings on the UNREAL-NH \cite{liu2023nighthazeformer}. The quantitative results demonstrate that the FDP, LP and BNM all have a positive effect on the dehazing performance. Since we must control the size of the paper, visualizations of the amplitude and phase spectrums are placed in Supplementary Materials.

\noindent
\textbf{Hierarchical Training and Frequency Domain Loss.} The training process of the proposed SFSNiD takes a hierarchical strategy by using differ scales $s \in \{0, 1, 2\}$. Two ablation studies are adopted, which are denoted as (i) $S1$: $s \in \{0\}$, and (ii) $S2$: $s \in \{0, 1\}$. Meanwhile, in our experimental setup, the spatial domain loss $L_{G}$ and the frequency domain loss $L_{F}$ are applied simultaneously. To verify the effectiveness of frequency domain loss, the setting when $L_{F}$ is not used is denoted as $S3$ ($s \in \{0, 1, 2\}$). Table \ref{tab:ablation_on_different_scale} shows the ablation results under the three different settings. The quantitative results demonstrate two main conclusions. First, the hierarchical training strategy can improve the dehazing performance. Second, the loss in the frequency domain is crucial as it improves the SSIM from 0.816 to 0.862.

\begin{figure}
  \footnotesize
  \centering
  \includegraphics[width=2.02cm,height=1.2cm]{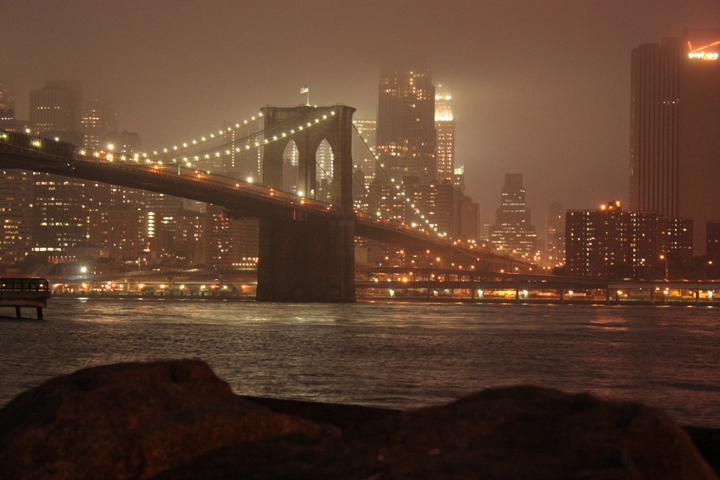}
  \includegraphics[width=2.02cm,height=1.2cm]{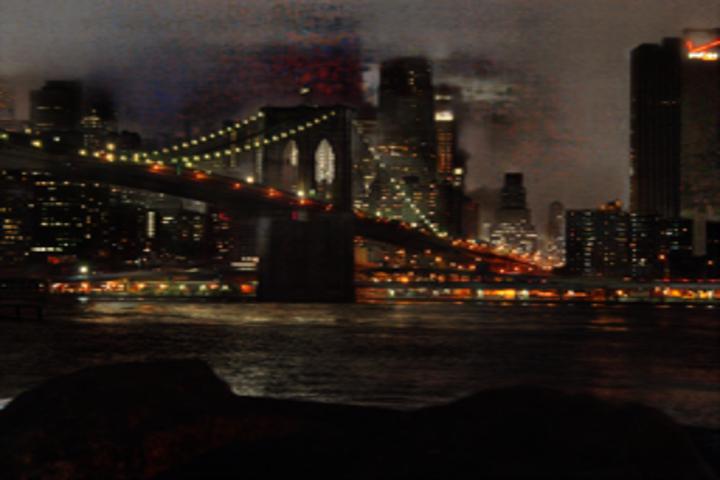}
  \includegraphics[width=2.02cm,height=1.2cm]{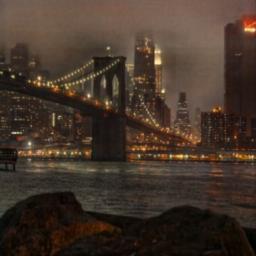}
  \includegraphics[width=2.02cm,height=1.2cm]{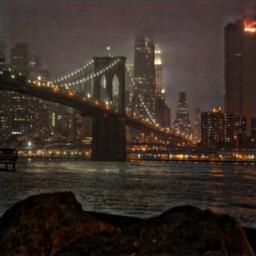}
  \\
  \includegraphics[width=2.02cm,height=1.2cm]{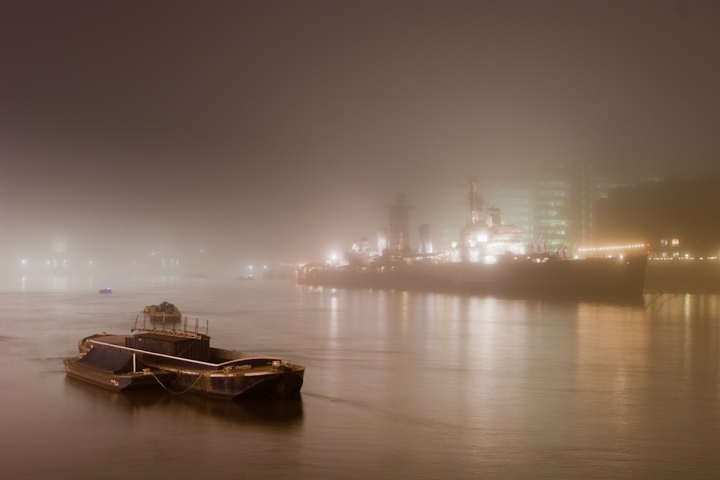}
  \includegraphics[width=2.02cm,height=1.2cm]{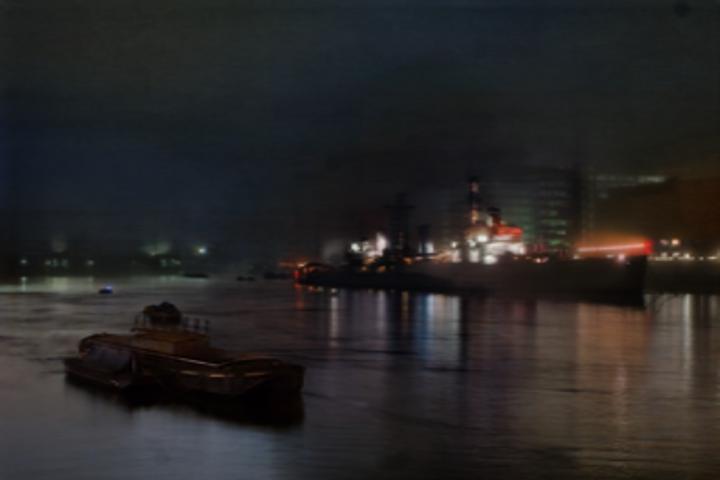}
  \includegraphics[width=2.02cm,height=1.2cm]{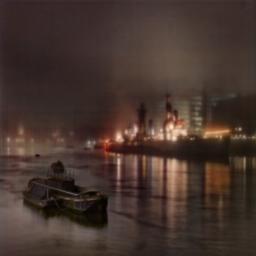}
  \includegraphics[width=2.02cm,height=1.2cm]{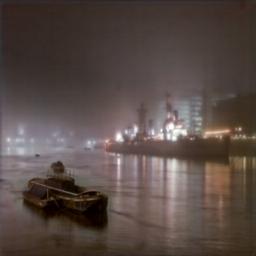}

  \leftline{\hspace{0.5cm} (a) Hazy \hspace{0.4cm} (b) Pseudo Label \hspace{0.25cm} (c) Retraining \hspace{0.1cm} (d) Retraining + $\mathcal{L}_{B}$ }

  \caption{Visual results under different training strategies.}
  \label{fig:ablation_before_and_after_retraining}
\end{figure}

\noindent
\textbf{Retraining Strategy and Realistic Brightness Loss.} To verify the effectiveness of the retraining strategy and the realistic brightness loss $\mathcal{L}_{B}$, the visual effects are shown in Figure \ref{fig:ablation_before_and_after_retraining}. As shown in Figure \ref{fig:ablation_before_and_after_retraining}-(\textcolor{red}{{b}}), the texture of the pseudo-labels is blurred due to the domain discrepancy between the synthetic and real-world data. The dehazed images obtained after retraining has unrealistic brightness as shown in Figure \ref{fig:ablation_before_and_after_retraining}-(\textcolor{red}{{c}}). It can be seen that the best effect occurs when the retraining strategy and $\mathcal{L}_{B}$ are used simultaneously as shown in Figure \ref{fig:ablation_before_and_after_retraining}-(\textcolor{red}{{d}}). The BRISQUE ($\downarrow$) and MUSIQ ($\uparrow$) obtained for the three settings (b), (c) and (d) in Figure \ref{fig:ablation_before_and_after_retraining} are $\{33.316, 30.432\}$, $\{34.210,32.373\}$ and $\{30.975,32.120\}$, respectively. Taking a comprehensive look at the visual and quantitative evaluation results, our proposed strategy is effective.


\begin{table}
  \scriptsize
  \centering
  \setlength{\tabcolsep}{1.5mm}
  \renewcommand{\arraystretch}{1.2}
  \caption{Quantitative results on datasets generated by game engine (GTA5 and UNREAL-NH) and the real-world dataset (RWNH).}
  \label{tab:nighttime_dehazing_results_game_engine}
  
  \begin{tabular}{cc@{\hspace{0.23cm}}cc@{\hspace{0.23cm}}cc@{\hspace{0.23cm}}c}
    \hline
      \multirow{2}{*}{Methods} & \multicolumn{2}{c}{GTA5} & \multicolumn{2}{c}{UNREAL-NH} & \multicolumn{2}{c}{RWNH} \\
    \cline{2-7}
     & SSIM$\uparrow$ & PSNR$\uparrow$ & SSIM$\uparrow$ & PSNR$\uparrow$ & BRISQUE $\downarrow$ & MUSIQ $\uparrow$ \\

    \hline   

    MRP          & 0.662 & 19.460 & 0.467 & 10.039 & \textbf{19.418} & 41.194 \\
    GD           & 0.900 & 30.090 & 0.767 & 21.202 & 31.359 & 33.433 \\
    OSFD         & 0.711 & 21.461 & 0.443 & 9.169  & 20.860 & \textbf{41.779} \\
    MSBDN        & 0.909 & 32.029 & 0.827 & 25.680 & 38.910 & 29.968 \\
    4KDehazing   & 0.903 & 30.314 & 0.774 & 23.087 & 34.965 & 33.536 \\
    AECRNet      & 0.888 & 26.846 & 0.731 & 21.566 & 27.084 & 37.034 \\
    DeHamer      & 0.928 & 32.597 & 0.740 & 22.441 & 42.269 & 26.788 \\
    FSDGN        & 0.923 & 32.642 & 0.702 & 21.736 & 32.216 & 35.200 \\
    DF           & 0.918 & 32.856 & 0.770 & 23.017 & 33.678 & 31.663 \\
    MITNet       & 0.899 & 31.118 & 0.766 & 21.860 & 35.404 & 31.768 \\
    Fourmer      & 0.917 & 31.926 & 0.772 & 22.799 & 35.850 & 31.367 \\
    Ours         & \textbf{0.935} & \textbf{33.708} & \textbf{0.862} & \textbf{25.907} & 30.975 & 32.120 \\

  \hline
  \end{tabular}
\end{table}

\begin{figure}
  \footnotesize
  \centering
  \includegraphics[width=8cm,height=3.0cm]{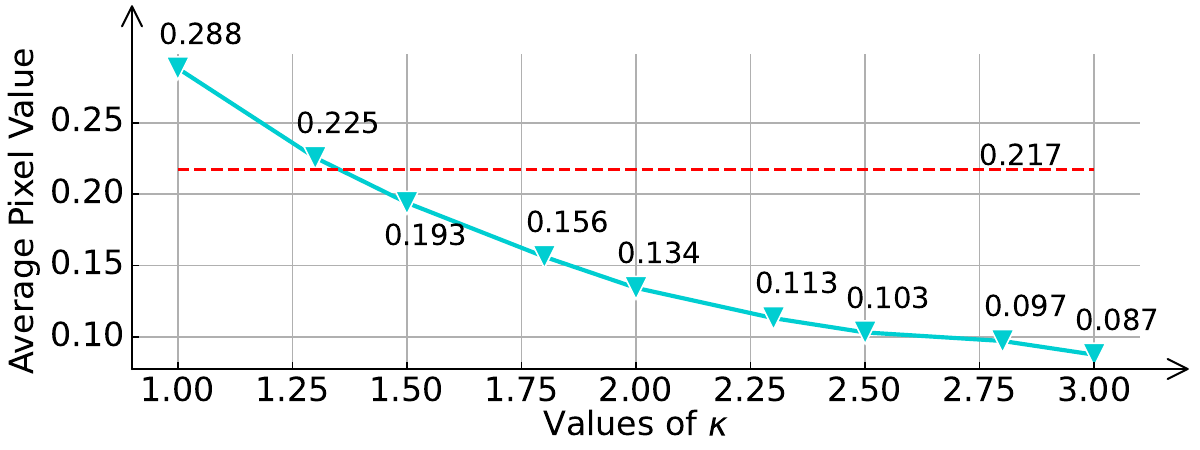}

  \caption{The average pixel value obtained under different $\kappa$. The horizontal dashed line represents the average pixel value of real-world nighttime clear images \cite{jin2023enhancing}.}
  \label{fig:obtained_brightness_ratio_when_brightness_ratio_is_set_to_different_values}
\end{figure}

\begin{table}
  \footnotesize
  \centering

  \setlength{\tabcolsep}{2.85mm}

  \caption{Ablation study on the SFII.}
  \label{tab:ablation_on_basic_block}
  
  \begin{tabular}{cccccc}
  \hline   
  Settings & $R1$ & $R2$ & $R3$ & $R4$ & Ours \\
  \hline
  SSIM     & 0.848  & 0.858  & 0.851  & 0.845  & \textbf{0.862} \\
  PSNR     & 25.353 & 25.808 & 25.642 & 24.301 & \textbf{25.907} \\
  \hline
  \end{tabular}
\end{table}

\begin{table}
  \footnotesize
  \centering

  \setlength{\tabcolsep}{4.0mm}

  \caption{Ablation study on the scale loss and frequency loss.}
  \label{tab:ablation_on_different_scale}
  
  \begin{tabular}{ccccc}
    \hline   
    Settings & $S1$ & $S2$ & $S3$ & Ours \\
    \hline
    SSIM     & 0.854  & 0.851  & 0.816  & \textbf{0.862} \\
    PSNR     & 25.601 & 25.134 & 24.464 & \textbf{25.907} \\
    \hline
  \end{tabular}

\end{table}

\noindent
\textbf{Brightness intensity coefficient $\kappa$ in $\mathcal{L}_{B}$.} In order to demonstrate the effectiveness of $\kappa$ on the real-world dehazing task, we manually set $\kappa$ to different values. The dehazed images and average pixel value when $\kappa$ takes different values are shown in Figure \ref{fig:visual_results_differ_brightness_intensity_coefficient_kappa} and Figure \ref{fig:obtained_brightness_ratio_when_brightness_ratio_is_set_to_different_values}, respectively. There are two conclusions that can be drawn. First, as $\kappa$ increases, the brightness of the dehazed image continues to decrease, which proves that $\kappa$ can control the brightness of the dehazed image. Second, when $\kappa$ equals $1.3$, the average pixel value ($0.225$) of dehazed images is close to the average pixel value real-world nighttime clear images ($0.217$) \cite{jin2023enhancing}. Therefore, we set $\kappa$ to $1.3$ as the final setting.


\section{Conclusion}
In this paper, a semi-supervised nighttime image dehazing baseline SFSNiD is proposed for real-world nighttime dehazing. A spatial and frequency domain information interaction module is proposed to handle the haze, glow, and noise with localized, coupled and frequency inconsistent characteristics. A retraining strategy and a local window-based brightness loss for semi-supervised training process are designed to suppress haze and glow while achieving realistic brightness. Experiments on public benchmarks validate the effectiveness of the proposed method and its superiority over state-of-the-art methods.

\textbf{Acknowledgment.} This work was supported in part by the grant of the National Science Foundation of China under Grant 62172090; Start-up Research Fund of Southeast University under Grant RF1028623097; CAAI-Huawei MindSpore Open Fund. We thank the Big Data Computing Center of Southeast University for providing the facility support on the numerical calculations in this paper.

\clearpage
\clearpage
{
    \small
    \bibliographystyle{ieeenat_fullname}
    \bibliography{main}

\begin{thebibliography}{56}
\providecommand{\natexlab}[1]{#1}
\providecommand{\url}[1]{\texttt{#1}}
\expandafter\ifx\csname urlstyle\endcsname\relax
  \providecommand{\doi}[1]{doi: #1}\else
  \providecommand{\doi}{doi: \begingroup \urlstyle{rm}\Url}\fi

\bibitem[Ancuti et~al.(2020)Ancuti, Ancuti, De~Vleeschouwer, and
  Bovik]{ancuti2020day}
Cosmin Ancuti, Codruta~O Ancuti, Christophe De~Vleeschouwer, and Alan~C Bovik.
\newblock Day and night-time dehazing by local airlight estimation.
\newblock \emph{IEEE Transactions on Image Processing}, 29:\penalty0
  6264--6275, 2020.

\bibitem[Cong et~al.(2020)Cong, Gui, Miao, Zhang, Wang, and
  Chen]{cong2020discrete}
Xiaofeng Cong, Jie Gui, Kai-Chao Miao, Jun Zhang, Bing Wang, and Peng Chen.
\newblock Discrete haze level dehazing network.
\newblock In \emph{ACM International Conference on Multimedia}, pages
  1828--1836, 2020.

\bibitem[Cui et~al.(2022)Cui, Tao, Bing, Ren, Gao, Cao, Huang, and
  Knoll]{cui2022selective}
Yuning Cui, Yi Tao, Zhenshan Bing, Wenqi Ren, Xinwei Gao, Xiaochun Cao, Kai
  Huang, and Alois Knoll.
\newblock Selective frequency network for image restoration.
\newblock In \emph{The Eleventh International Conference on Learning
  Representations}, 2022.

\bibitem[Dai et~al.(2022)Dai, Li, Zhou, Feng, and Loy]{dai2022flare7k}
Yuekun Dai, Chongyi Li, Shangchen Zhou, Ruicheng Feng, and Chen~Change Loy.
\newblock Flare7k: A phenomenological nighttime flare removal dataset.
\newblock \emph{Advances in Neural Information Processing Systems},
  35:\penalty0 3926--3937, 2022.

\bibitem[Dong et~al.(2020)Dong, Pan, Xiang, Hu, Zhang, Wang, and
  Yang]{dong2020multi}
Hang Dong, Jinshan Pan, Lei Xiang, Zhe Hu, Xinyi Zhang, Fei Wang, and
  Ming-Hsuan Yang.
\newblock Multi-scale boosted dehazing network with dense feature fusion.
\newblock In \emph{IEEE Conference on Computer Vision and Pattern Recognition},
  pages 2157--2167, 2020.

\bibitem[Frigo and Johnson(1998)]{frigo1998fftw}
Matteo Frigo and Steven~G Johnson.
\newblock Fftw: An adaptive software architecture for the fft.
\newblock In \emph{IEEE International Conference on Acoustics, Speech and
  Signal Processing}, pages 1381--1384, 1998.

\bibitem[Gui et~al.(2023)Gui, Cong, Cao, Ren, Zhang, Zhang, Cao, and
  Tao]{gui2023a}
Jie Gui, Xiaofeng Cong, Yuan Cao, Wenqi Ren, Jun Zhang, Jing Zhang, Jiuxin Cao,
  and Dacheng Tao.
\newblock A comprehensive survey and taxonomy on single image dehazing based on
  deep learning.
\newblock \emph{ACM Computing Surveys}, 2023.

\bibitem[Guo et~al.(2020)Guo, Li, Guo, Loy, Hou, Kwong, and Cong]{guo2020zero}
Chunle Guo, Chongyi Li, Jichang Guo, Chen~Change Loy, Junhui Hou, Sam Kwong,
  and Runmin Cong.
\newblock Zero-reference deep curve estimation for low-light image enhancement.
\newblock In \emph{IEEE Conference on Computer Vision and Pattern Recognition},
  pages 1780--1789, 2020.

\bibitem[Guo et~al.(2022{\natexlab{a}})Guo, Yan, Anwar, Cong, Ren, and
  Li]{guo2022image}
Chun-Le Guo, Qixin Yan, Saeed Anwar, Runmin Cong, Wenqi Ren, and Chongyi Li.
\newblock Image dehazing transformer with transmission-aware 3d position
  embedding.
\newblock In \emph{IEEE Conference on Computer Vision and Pattern Recognition},
  pages 5812--5820, 2022{\natexlab{a}}.

\bibitem[Guo et~al.(2022{\natexlab{b}})Guo, Fu, Zhou, Huang, Peng, and
  Zha]{guo2022exploring}
Xin Guo, Xueyang Fu, Man Zhou, Zhen Huang, Jialun Peng, and Zheng-Jun Zha.
\newblock Exploring fourier prior for single image rain removal.
\newblock In \emph{International Joint Conferences on Artificial Intelligence},
  pages 935--941, 2022{\natexlab{b}}.

\bibitem[Hou et~al.(2023)Hou, Cao, Ran, Liu, Li, and Deng]{hou2023bidomain}
Junming Hou, Qi Cao, Ran Ran, Che Liu, Junling Li, and Liang-jian Deng.
\newblock Bidomain modeling paradigm for pansharpening.
\newblock In \emph{ACM International Conference on Multimedia}, pages 347--357,
  2023.

\bibitem[Hu et~al.(2018)Hu, Shen, and Sun]{hu2018squeeze}
Jie Hu, Li Shen, and Gang Sun.
\newblock Squeeze-and-excitation networks.
\newblock In \emph{IEEE Conference on Computer Vision and Pattern Recognition},
  pages 7132--7141, 2018.

\bibitem[Huang et~al.(2023)Huang, Wang, Liu, Chen, and
  Li]{huang2023contrastive}
Shirui Huang, Keyan Wang, Huan Liu, Jun Chen, and Yunsong Li.
\newblock Contrastive semi-supervised learning for underwater image restoration
  via reliable bank.
\newblock In \emph{IEEE Conference on Computer Vision and Pattern Recognition},
  pages 18145--18155, 2023.

\bibitem[Jin et~al.(2023)Jin, Lin, Yan, Ye, Yuan, and Tan]{jin2023enhancing}
Yeying Jin, Beibei Lin, Wending Yan, Wei Ye, Yuan Yuan, and Robby~T Tan.
\newblock Enhancing visibility in nighttime haze images using guided apsf and
  gradient adaptive convolution.
\newblock In \emph{ACM International Conference on Multimedia}, 2023.

\bibitem[Ju et~al.(2021{\natexlab{a}})Ju, Ding, Guo, Ren, and Tao]{ju2021idrlp}
Mingye Ju, Can Ding, Charles~A Guo, Wenqi Ren, and Dacheng Tao.
\newblock Idrlp: Image dehazing using region line prior.
\newblock \emph{IEEE Transactions on Image Processing}, 30:\penalty0
  9043--9057, 2021{\natexlab{a}}.

\bibitem[Ju et~al.(2021{\natexlab{b}})Ju, Ding, Ren, Yang, Zhang, and
  Guo]{ju2021ide}
Mingye Ju, Can Ding, Wenqi Ren, Yi Yang, Dengyin Zhang, and Y~Jay Guo.
\newblock Ide: Image dehazing and exposure using an enhanced atmospheric
  scattering model.
\newblock \emph{IEEE Transactions on Image Processing}, 30:\penalty0
  2180--2192, 2021{\natexlab{b}}.

\bibitem[Ke et~al.(2021)Ke, Wang, Wang, Milanfar, and Yang]{ke2021musiq}
Junjie Ke, Qifei Wang, Yilin Wang, Peyman Milanfar, and Feng Yang.
\newblock Musiq: Multi-scale image quality transformer.
\newblock In \emph{IEEE International Conference on Computer Vision}, pages
  5148--5157, 2021.

\bibitem[Koo and Kim(2020)]{koo2020nighttime}
Beomhyuk Koo and Gyeonghwan Kim.
\newblock Nighttime haze removal with glow decomposition using gan.
\newblock In \emph{Pattern Recognition: 5th Asian Conference}, pages 807--820,
  2020.

\bibitem[Kuanar et~al.(2022)Kuanar, Mahapatra, Bilas, and Rao]{kuanar2022multi}
Shiba Kuanar, Dwarikanath Mahapatra, Monalisa Bilas, and KR Rao.
\newblock Multi-path dilated convolution network for haze and glow removal in
  nighttime images.
\newblock \emph{The Visual Computer}, pages 1--14, 2022.

\bibitem[Li et~al.(2017)Li, Peng, Wang, Xu, and Feng]{li2017aod}
Boyi Li, Xiulian Peng, Zhangyang Wang, Jizheng Xu, and Dan Feng.
\newblock Aod-net: All-in-one dehazing network.
\newblock In \emph{IEEE International Conference on Computer Vision}, pages
  4770--4778, 2017.

\bibitem[Li et~al.(2018)Li, Ren, Fu, Tao, Feng, Zeng, and
  Wang]{li2018benchmarking}
Boyi Li, Wenqi Ren, Dengpan Fu, Dacheng Tao, Dan Feng, Wenjun Zeng, and
  Zhangyang Wang.
\newblock Benchmarking single-image dehazing and beyond.
\newblock \emph{IEEE Transactions on Image Processing}, 28\penalty0
  (1):\penalty0 492--505, 2018.

\bibitem[Li et~al.(2023)Li, Guo, Zhou, Liang, Zhou, Feng, and
  Loy]{li2023embedding}
Chongyi Li, Chun-Le Guo, Man Zhou, Zhexin Liang, Shangchen Zhou, Ruicheng Feng,
  and Chen~Change Loy.
\newblock Embedding fourier for ultra-high-definition low-light image
  enhancement.
\newblock \emph{arXiv preprint arXiv:2302.11831}, 2023.

\bibitem[Li et~al.(2021)Li, Guo, and Wang]{li2021proposal}
Kun Li, Dan Guo, and Meng Wang.
\newblock Proposal-free video grounding with contextual pyramid network.
\newblock In \emph{AAAI Conference on Artificial Intelligence}, pages
  1902--1910, 2021.

\bibitem[Li et~al.(2015)Li, Tan, and Brown]{li2015nighttime}
Yu Li, Robby~T Tan, and Michael~S Brown.
\newblock Nighttime haze removal with glow and multiple light colors.
\newblock In \emph{IEEE International Conference on Computer Vision}, pages
  226--234, 2015.

\bibitem[Liang et~al.(2022)Liang, Wang, Zuo, Liu, and Ren]{liang2022self}
Yudong Liang, Bin Wang, Wangmeng Zuo, Jiaying Liu, and Wenqi Ren.
\newblock Self-supervised learning and adaptation for single image dehazing.
\newblock In \emph{International Joint Conference on Artificial Intelligence},
  pages 1--15, 2022.

\bibitem[Liao et~al.(2018)Liao, Su, Liang, and Qiu]{liao2018hdp}
Yinghong Liao, Zhuo Su, Xiangguo Liang, and Bin Qiu.
\newblock Hdp-net: Haze density prediction network for nighttime dehazing.
\newblock In \emph{Pacific Rim Conference on Multimedia}, pages 469--480, 2018.

\bibitem[Liu et~al.(2019)Liu, Ma, Shi, and Chen]{liu2019griddehazenet}
Xiaohong Liu, Yongrui Ma, Zhihao Shi, and Jun Chen.
\newblock Griddehazenet: Attention-based multi-scale network for image
  dehazing.
\newblock In \emph{IEEE International Conference on Computer Vision}, pages
  7314--7323, 2019.

\bibitem[Liu et~al.(2021{\natexlab{a}})Liu, Wang, Zhou, and Jia]{liu2021single}
Yun Liu, Anzhi Wang, Hao Zhou, and Pengfei Jia.
\newblock Single nighttime image dehazing based on image decomposition.
\newblock \emph{Signal Processing}, 183:\penalty0 107986, 2021{\natexlab{a}}.

\bibitem[Liu et~al.(2022{\natexlab{a}})Liu, Yan, Tan, and Li]{liu2022multi}
Yun Liu, Zhongsheng Yan, Jinge Tan, and Yuche Li.
\newblock Multi-purpose oriented single nighttime image haze removal based on
  unified variational retinex model.
\newblock \emph{IEEE Transactions on Circuits and Systems for Video
  Technology}, 33\penalty0 (4):\penalty0 1643--1657, 2022{\natexlab{a}}.

\bibitem[Liu et~al.(2022{\natexlab{b}})Liu, Yan, Wu, and Ye]{liu2022nighttime}
Yun Liu, Zhongsheng Yan, Aimin Wu, and Tian Ye.
\newblock Nighttime image dehazing based on variational decomposition model.
\newblock In \emph{IEEE Conference on Computer Vision and Pattern Recognition
  Workshops}, pages 640--649, 2022{\natexlab{b}}.

\bibitem[Liu et~al.(2023)Liu, Yan, Chen, Ye, Ren, and
  Chen]{liu2023nighthazeformer}
Yun Liu, Zhongsheng Yan, Sixiang Chen, Tian Ye, Wenqi Ren, and Erkang Chen.
\newblock Nighthazeformer: Single nighttime haze removal using prior query
  transformer.
\newblock In \emph{ACM International Conference on Multimedia}, 2023.

\bibitem[Liu et~al.(2021{\natexlab{b}})Liu, Lin, Cao, Hu, Wei, Zhang, Lin, and
  Guo]{liu2021swin}
Ze Liu, Yutong Lin, Yue Cao, Han Hu, Yixuan Wei, Zheng Zhang, Stephen Lin, and
  Baining Guo.
\newblock Swin transformer: Hierarchical vision transformer using shifted
  windows.
\newblock In \emph{IEEE International Conference on Computer Vision}, pages
  10012--10022, 2021{\natexlab{b}}.

\bibitem[Ren et~al.(2019)Ren, Liu, Ma, Xu, Xu, Cao, Du, and Yang]{ren2019low}
Wenqi Ren, Sifei Liu, Lin Ma, Qianqian Xu, Xiangyu Xu, Xiaochun Cao, Junping
  Du, and Ming-Hsuan Yang.
\newblock Low-light image enhancement via a deep hybrid network.
\newblock \emph{IEEE Transactions on Image Processing}, 28\penalty0
  (9):\penalty0 4364--4375, 2019.

\bibitem[Ren et~al.(2020)Ren, Pan, Zhang, Cao, and Yang]{ren2020single}
Wenqi Ren, Jinshan Pan, Hua Zhang, Xiaochun Cao, and Ming-Hsuan Yang.
\newblock Single image dehazing via multi-scale convolutional neural networks
  with holistic edges.
\newblock \emph{International Journal of Computer Vision}, 128:\penalty0
  240--259, 2020.

\bibitem[Shao et~al.(2020)Shao, Li, Ren, Gao, and Sang]{shao2020domain}
Yuanjie Shao, Lerenhan Li, Wenqi Ren, Changxin Gao, and Nong Sang.
\newblock Domain adaptation for image dehazing.
\newblock In \emph{IEEE Conference on Computer Vision and Pattern Recognition},
  pages 2808--2817, 2020.

\bibitem[Shen et~al.(2023{\natexlab{a}})Shen, Zhao, and
  Zhang]{shen2023adaptive}
Hao Shen, Zhong-Qiu Zhao, and Wandi Zhang.
\newblock Adaptive dynamic filtering network for image denoising.
\newblock In \emph{AAAI Conference on Artificial Intelligence}, pages
  2227--2235, 2023{\natexlab{a}}.

\bibitem[Shen et~al.(2023{\natexlab{b}})Shen, Zhao, Zhang, and
  Zhang]{shen2023mutual}
Hao Shen, Zhong-Qiu Zhao, Yulun Zhang, and Zhao Zhang.
\newblock Mutual information-driven triple interaction network for efficient
  image dehazing.
\newblock In \emph{ACM International Conference on Multimedia}, pages 7--16,
  2023{\natexlab{b}}.

\bibitem[Song et~al.(2023)Song, He, Qian, and Du]{song2023vision}
Yuda Song, Zhuqing He, Hui Qian, and Xin Du.
\newblock Vision transformers for single image dehazing.
\newblock \emph{IEEE TIP}, 32:\penalty0 1927--1941, 2023.

\bibitem[Sun et~al.(2022)Sun, Ren, and Wang]{sun2022rethinking}
Shangquan Sun, Wenqi Ren, and Tao Wang.
\newblock Rethinking image restoration for object detection.
\newblock \emph{Advances in Neural Information Processing Systems},
  35:\penalty0 4461--4474, 2022.

\bibitem[Wang et~al.(2023)Wang, Guo, and Li]{wang2023eulermormer}
Fei Wang, Dan Guo, and Kun Li.
\newblock Eulermormer: Robust eulerian motion magnification via dynamic
  filtering within transformer.
\newblock \emph{arXiv preprint arXiv:2312.04152}, 2023.

\bibitem[Wang et~al.(2022)Wang, Wang, and Liu]{wang2022variational}
Wenhui Wang, Anna Wang, and Chen Liu.
\newblock Variational single nighttime image haze removal with a gray haze-line
  prior.
\newblock \emph{IEEE Transactions on Image Processing}, 31:\penalty0
  1349--1363, 2022.

\bibitem[Wang et~al.(2004)Wang, Bovik, Sheikh, and Simoncelli]{wang2004image}
Zhou Wang, Alan~C Bovik, Hamid~R Sheikh, and Eero~P Simoncelli.
\newblock Image quality assessment: from error visibility to structural
  similarity.
\newblock \emph{IEEE Transactions on Image Processing}, 13\penalty0
  (4):\penalty0 600--612, 2004.

\bibitem[Wu et~al.(2021)Wu, Qu, Lin, Zhou, Qiao, Zhang, Xie, and
  Ma]{wu2021contrastive}
Haiyan Wu, Yanyun Qu, Shaohui Lin, Jian Zhou, Ruizhi Qiao, Zhizhong Zhang, Yuan
  Xie, and Lizhuang Ma.
\newblock Contrastive learning for compact single image dehazing.
\newblock In \emph{IEEE Conference on Computer Vision and Pattern Recognition},
  pages 10551--10560, 2021.

\bibitem[Wu et~al.(2023)Wu, Duan, Guo, Chai, and Li]{wu2023ridcp}
Rui-Qi Wu, Zheng-Peng Duan, Chun-Le Guo, Zhi Chai, and Chongyi Li.
\newblock Ridcp: Revitalizing real image dehazing via high-quality codebook
  priors.
\newblock In \emph{IEEE Conference on Computer Vision and Pattern Recognition},
  pages 22282--22291, 2023.

\bibitem[Yan et~al.(2020)Yan, Tan, and Dai]{yan2020nighttime}
Wending Yan, Robby~T Tan, and Dengxin Dai.
\newblock Nighttime defogging using high-low frequency decomposition and
  grayscale-color networks.
\newblock In \emph{European Conference on Computer Vision}, pages 473--488,
  2020.

\bibitem[Yang et~al.(2018)Yang, Liu, and Li]{yang2018superpixel}
Minmin Yang, Jianchang Liu, and Zhengguo Li.
\newblock Superpixel-based single nighttime image haze removal.
\newblock \emph{IEEE Transactions on Multimedia}, 20\penalty0 (11):\penalty0
  3008--3018, 2018.

\bibitem[Yu et~al.(2022)Yu, Zheng, Zhou, Huang, Xiao, and
  Zhao]{yu2022frequency}
Hu Yu, Naishan Zheng, Man Zhou, Jie Huang, Zeyu Xiao, and Feng Zhao.
\newblock Frequency and spatial dual guidance for image dehazing.
\newblock In \emph{European Conference on Computer Vision}, pages 181--198,
  2022.

\bibitem[Zhang and Tao(2019)]{zhang2019famed}
Jing Zhang and Dacheng Tao.
\newblock Famed-net: A fast and accurate multi-scale end-to-end dehazing
  network.
\newblock \emph{IEEE Transactions on Image Processing}, 29:\penalty0 72--84,
  2019.

\bibitem[Zhang et~al.(2014)Zhang, Cao, and Wang]{zhang2014nighttime}
Jing Zhang, Yang Cao, and Zengfu Wang.
\newblock Nighttime haze removal based on a new imaging model.
\newblock In \emph{IEEE International Conference on Image Processing}, pages
  4557--4561, 2014.

\bibitem[Zhang et~al.(2017)Zhang, Cao, Fang, Kang, and Wen~Chen]{zhang2017fast}
Jing Zhang, Yang Cao, Shuai Fang, Yu Kang, and Chang Wen~Chen.
\newblock Fast haze removal for nighttime image using maximum reflectance
  prior.
\newblock In \emph{IEEE Conference on Computer Vision and Pattern Recognition},
  pages 7418--7426, 2017.

\bibitem[Zhang et~al.(2020)Zhang, Cao, Zha, and Tao]{zhang2020nighttime}
Jing Zhang, Yang Cao, Zheng-Jun Zha, and Dacheng Tao.
\newblock Nighttime dehazing with a synthetic benchmark.
\newblock In \emph{ACM International Conference on Multimedia}, pages
  2355--2363, 2020.

\bibitem[Zhang et~al.(2021{\natexlab{a}})Zhang, Ren, Zhang, Zhang, Nie, Xue,
  and Cao]{zhang2021hierarchical}
Jingang Zhang, Wenqi Ren, Shengdong Zhang, He Zhang, Yunfeng Nie, Zhe Xue, and
  Xiaochun Cao.
\newblock Hierarchical density-aware dehazing network.
\newblock \emph{IEEE Transactions on Cybernetics}, 52\penalty0 (10):\penalty0
  11187--11199, 2021{\natexlab{a}}.

\bibitem[Zhang et~al.(2021{\natexlab{b}})Zhang, Ren, Tan, Wang, Liu, Zhang,
  Zhang, and Cao]{zhang2021semantic}
Shengdong Zhang, Wenqi Ren, Xin Tan, Zhi-Jie Wang, Yong Liu, Jingang Zhang,
  Xiaoqin Zhang, and Xiaochun Cao.
\newblock Semantic-aware dehazing network with adaptive feature fusion.
\newblock \emph{IEEE Transactions on Cybernetics}, 53\penalty0 (1):\penalty0
  454--467, 2021{\natexlab{b}}.

\bibitem[Zheng et~al.(2021)Zheng, Ren, Cao, Hu, Wang, Song, and
  Jia]{zheng2021ultra}
Zhuoran Zheng, Wenqi Ren, Xiaochun Cao, Xiaobin Hu, Tao Wang, Fenglong Song,
  and Xiuyi Jia.
\newblock Ultra-high-definition image dehazing via multi-guided bilateral
  learning.
\newblock In \emph{IEEE Conference on Computer Vision and Pattern Recognition},
  pages 16180--16189, 2021.

\bibitem[Zhou et~al.(2023{\natexlab{a}})Zhou, Huang, Guo, and
  Li]{zhou2023fourmer}
Man Zhou, Jie Huang, Chun-Le Guo, and Chongyi Li.
\newblock Fourmer: an efficient global modeling paradigm for image restoration.
\newblock In \emph{International Conference on Machine Learning}, pages
  42589--42601, 2023{\natexlab{a}}.

\bibitem[Zhou et~al.(2023{\natexlab{b}})Zhou, Yan, Fu, Liu, and
  Xie]{zhou2023pan}
Man Zhou, Keyu Yan, Xueyang Fu, Aiping Liu, and Chengjun Xie.
\newblock Pan-guided band-aware multi-spectral feature enhancement for
  pan-sharpening.
\newblock \emph{IEEE Transactions on Computational Imaging}, 9:\penalty0
  238--249, 2023{\natexlab{b}}.

\end{thebibliography}
}

\end{document}